\documentclass[lettersize,journal]{IEEEtran}
\usepackage{cite}
\usepackage{amsmath,amssymb,amsfonts}
\usepackage{bm}
\usepackage{algorithmic}
\usepackage{graphicx}
\usepackage{pifont}
\usepackage{textcomp}
\usepackage{hyperref}
\usepackage{multirow}
\usepackage{balance}
\usepackage[table,xcdraw]{xcolor}
\usepackage{url}
\usepackage{comment}
\usepackage{color}
\usepackage{latexsym}
\usepackage{mathtools}
\usepackage{caption}
\usepackage{subcaption}
\usepackage{orcidlink}
\newcommand{\cmark}{\ding{51}} 
\newcommand{\xmark}{\ding{55}} 
\newcommand{\smark}{\ding{72}} 
\begin{document}

\title{
When Digital Twins Meet Large Language Models: Realistic,\\Interactive, and Editable Simulation for Autonomous Driving
}

\author{Tanmay Samak$^{\star \dagger}$ \orcidlink{0000-0002-9717-0764}, Chinmay Samak$^{\star \dagger}$ \orcidlink{0000-0002-6455-6716}, Bing Li$^{\dagger}$ \orcidlink{0000-0003-4987-6129}, and Venkat Krovi$^{\dagger}$ \orcidlink{0000-0003-2539-896X}
\thanks{$^{\star}$These authors contributed equally.}
\thanks{$^{\dagger}$Department of Automotive Engineering, Clemson University International Center for Automotive Research (CU-ICAR), Greenville, SC 29607, USA.
{\tt\small {\{\href{mailto:tsamak@clemson.edu}{tsamak}, \href{mailto:csamak@clemson.edu}{csamak}, \href{mailto:bli4@clemson.edu}{bli4}, \href{mailto:vkrovi@clemson.edu}{vkrovi}\}@clemson.edu}}}%
}



\maketitle

\begin{abstract}
Simulation frameworks have been key enablers for the development and validation of autonomous driving systems. However, existing methods struggle to comprehensively address the autonomy-oriented requirements of balancing: (i) dynamical fidelity, (ii) photorealistic rendering, (iii) context-relevant scenario orchestration, and (iv) real-time performance. To address these limitations, we present a unified framework for creating and curating high-fidelity digital twins to accelerate advancements in autonomous driving research. Our framework leverages a mix of physics-based and data-driven techniques for developing and simulating digital twins of autonomous vehicles and their operating environments. It is capable of reconstructing real-world scenes and assets with geometric and photorealistic accuracy ($\sim$97\% structural similarity) and infusing them with physical properties to enable real-time ($>$60 Hz) dynamical simulation of the ensuing driving scenarios. Additionally, it incorporates a large language model (LLM) interface to flexibly edit the driving scenarios online via natural language prompts, with $\sim$85\% generalizability and $\sim$95\% repeatability. Finally, an optional vision language model (VLM) provides $\sim$80\% visual enhancement by blending the hybrid scene composition.\\%
\end{abstract}

\begin{IEEEkeywords}
Agentic AI, Generative AI, Spatial AI, Scene Reconstruction, Scenario Generation, Autonomous Vehicles, Digital Twins, Real2Sim Transfer\\%
\end{IEEEkeywords}


\section{Introduction}
\label{Section: Introduction}

\begin{figure}[t]
     \centering
     \includegraphics[width=\linewidth]{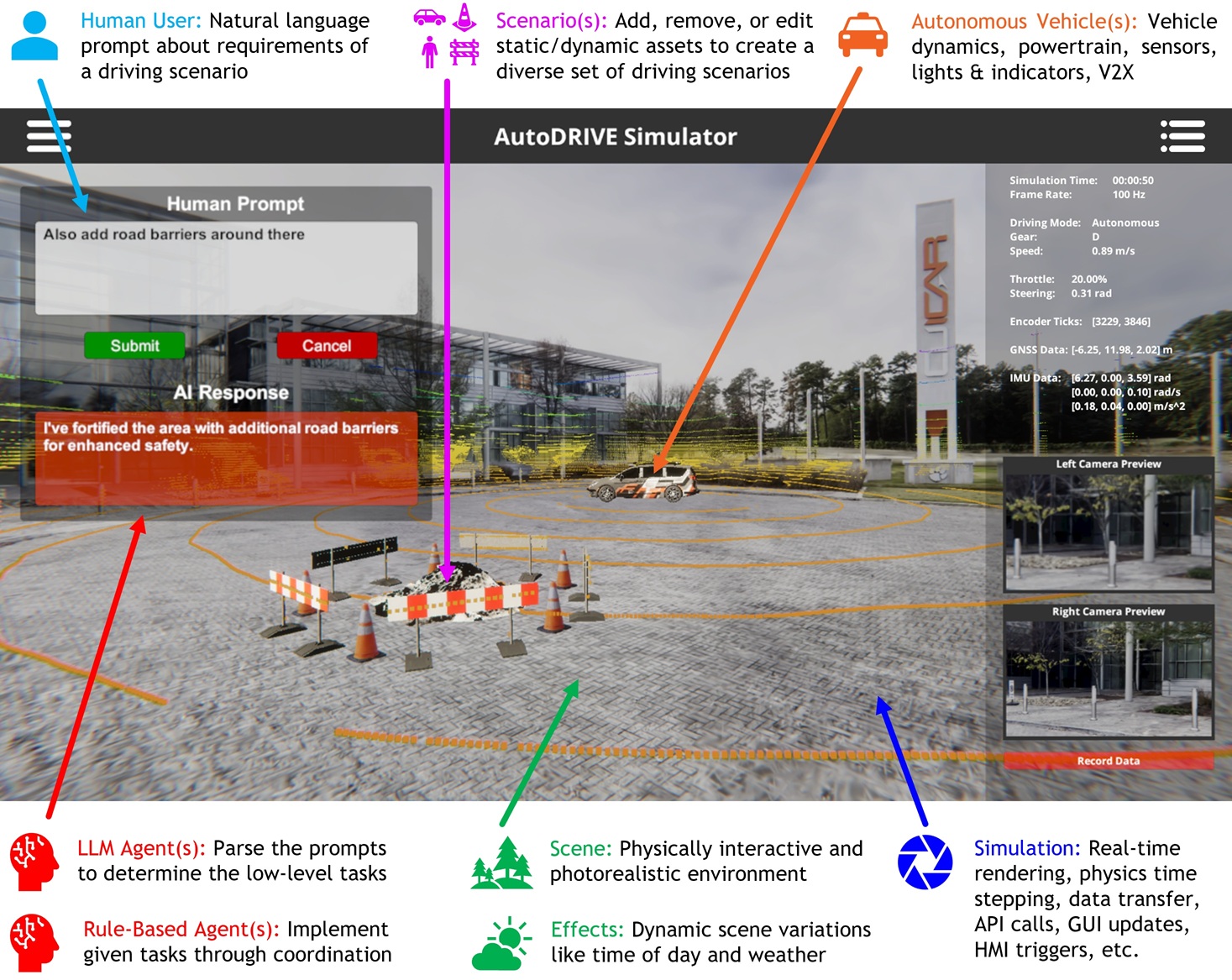}
     \caption{Proposed framework for curating autonomy-oriented digital twins, which blends photorealism and physical interaction with LLM-driven scenario orchestration for enhanced serviceability. \textbf{Video:} \url{https://youtu.be/3ovFiRgbHFc}}
     \label{fig1}
\end{figure}

\IEEEPARstart{T}{he} development and deployment of autonomous driving systems requires extensive training/optimization and testing/validation in a wide variety of scenarios to ensure safety, robustness, and scalability. Real-world training/testing is often limited by high costs, large time investments, safety concerns, and the lack of control over creating edge cases or extreme conditions. In this context, digital twins offer a tantalizing alternative by providing controllable, repeatable, and diverse simulations for accelerating synthetic data generation, design optimization as well as systematic verification and validation. However, creating high-fidelity digital twins that simultaneously balance photorealism, physical accuracy, operational flexibility, and real-time performance remains a significant challenge -- one which has typically been underexplored in the literature.

To this end, we propose a unified framework (refer Fig. \ref{fig1}) for reconstructing high-fidelity digital twins of autonomous vehicles and their operating environments, and reconfiguring them via artificial intelligence (AI) agents. Our framework is designed to meet the specific requirements of autonomy-oriented digital twins: (a) physical accuracy and interactivity to simulate vehicle dynamics, sensor characteristics, and environment physics; (b) geometric and visual fidelity to support feature-rich and reliable exteroceptive perception, (c) contextual understanding for generating diverse and creative driving scenarios, and (d) real-time interfacing with autonomous driving software stacks.

\begin{figure*}[t]
     \centering
     \includegraphics[width=\linewidth]{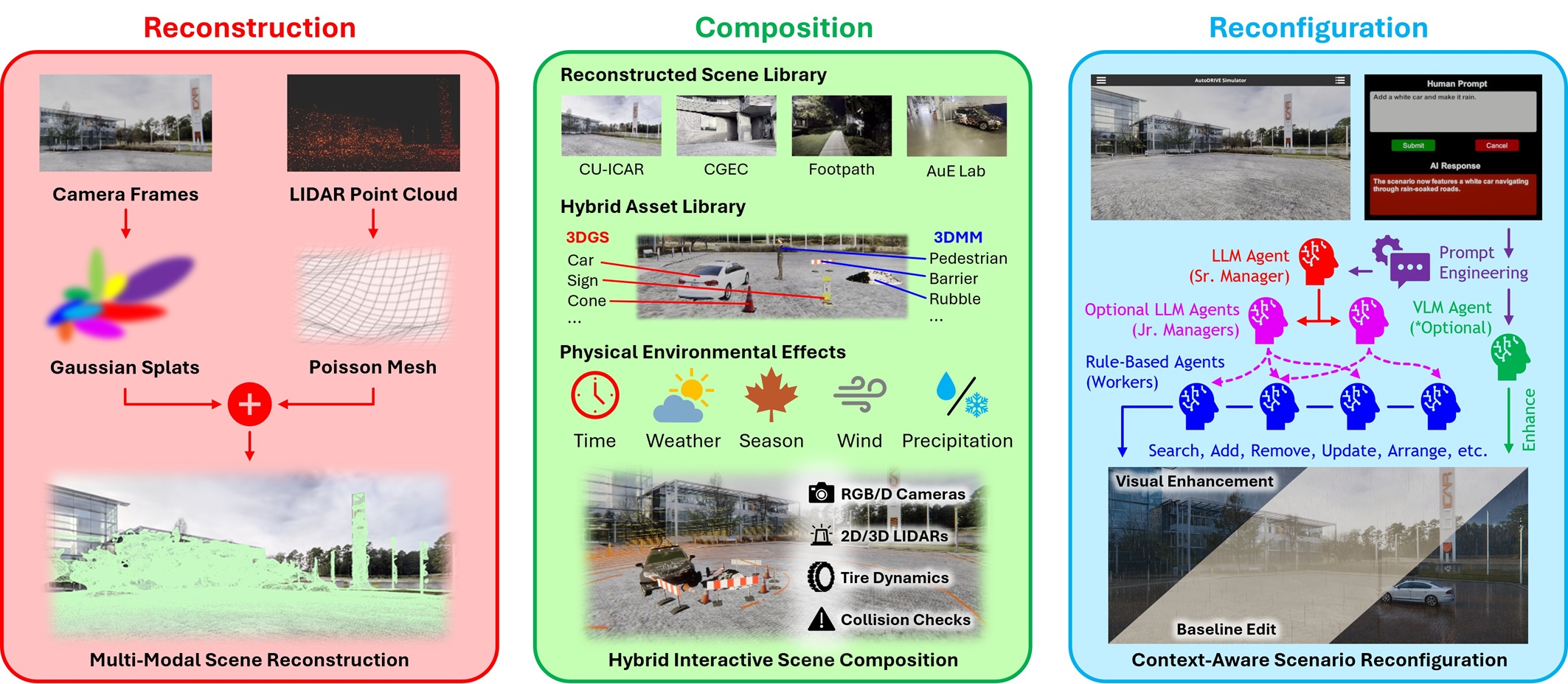}
     \caption{Proposed approach to (a) photorealistic and geometrically accurate 3D scene reconstruction, (b) hybrid scene composition of physics-based and data-driven digital twins for physically interactive and graphically realistic simulation, and (c) LLM-guided context-aware scenario reconfiguration with optional VLM-guided visual enhancement.}
     \label{fig2}
\end{figure*}

The core novelty of our framework lies in its ability to reconstruct real-world scenes and assets with geometric and photorealistic accuracy, and to seamlessly blend them with synthetic 3D elements represented as point, surface, or mesh models, through a real-time hybrid physics and rendering engine. By infusing physical properties into these assets, the framework enables real-time, dynamic simulation of driving scenarios that capture realistic interactions among vehicles, pedestrians, and environmental elements. To further enhance its serviceability, the framework integrates an intuitive human-machine interface (HMI) powered by large language models (LLMs), enabling users to perform creative and interesting edits to generate a diverse range of driving scenarios in real time via natural language prompts. Finally, an optional vision language model (VLM) provides appearance enhancement by blending the hybrid scene composition. Unlike existing solutions that often focus on isolated aspects like physics (e.g., vehicle dynamics), graphics (e.g., static scene rendering), or context (e.g., predefined scenarios), our framework offers a comprehensive reality-to-simulation (real2sim) workflow with real-time performance. This holistic approach not only enhances simulation fidelity but also enables convenient prototyping of autonomous driving systems.

The key contributions of this paper are summarized below:

\begin{itemize}
    \item \textbf{High-Fidelity Reconstruction:} We propose a multi-modal method to reconstruct real-world scenes/assets from camera and LIDAR data and seamlessly create visually, geometrically, and physically accurate, interactive simulations.
    \item \textbf{Automated Reconfiguration:} We present a hierarchical method comprising a mix of LLM-driven and rule-based agent(s) to automatically generate driving scenarios using natural language prompts. The proposed method fosters the creativity of generative AI, while enforcing rule-based bounds to generate diverse yet pragmatic scenarios.
    \item \textbf{Optional Enhancement:} We introduce an optional VLM to enhance the raw rendering into a more \textit{``realistic''} one, while imposing explicit constraints on modifying the pose, scale, or other attributes of the scene/assets.
    \item \textbf{Experimental Evaluation:} We experiment with the reconstruction and reconfiguration techniques to analyze their accuracy, performance, repeatability, and generalizability. We also present qualitative insights in terms of data expectations and prompt engineering.
    \item \textbf{Open-Source Contribution:} We openly release the proposed framework as a part of AutoDRIVE Ecosystem\footnote{\textbf{AutoDRIVE:} \url{https://autodrive-ecosystem.github.io}} \cite{AutoDRIVE2023}, to enable further research in the field of autonomous driving and digital twins.
\end{itemize}

The remainder of this paper is structured as follows: Section \ref{Section: Related Work} discusses related work and positions our framework within the existing landscape. Section \ref{Section: Research Methodology} details the architecture and implementation of the framework, including its core subsystems. Section \ref{Section: Results and Discussion} presents experimental results and performance benchmarks. Finally, Section \ref{Section: Conclusion} concludes with key insights and future directions.


\section{Related Work}
\label{Section: Related Work}

Simulation frameworks play a crucial role in the field of autonomous driving, as they alleviate the financial, safety, spatial, and temporal constraints imposed during physical development and validation. However, state-of-the-art frameworks struggle to deliver a unified solution that combines truly photorealistic rendering with physically accurate, real-time simulations. This includes the system-of-systems level simulation of vehicles (sensors, actuators, vehicle dynamics), environments (structure, appearance, variability), and their complex interactions. Additionally, most of the existing approaches lack efficient user interfaces to orchestrate context-aware scenarios that capture the diverse nature of real-world driving conditions. Below, we review some of the related works, which can be broadly categorized based on their isolated focus on scene reconstruction, scenario reconfiguration, or autonomy-oriented simulation.

\subsection{Scene Reconstruction}

\begin{figure*}[t]
     \centering
     \begin{subfigure}[b]{0.329\linewidth}
         \centering
         \includegraphics[width=\linewidth]{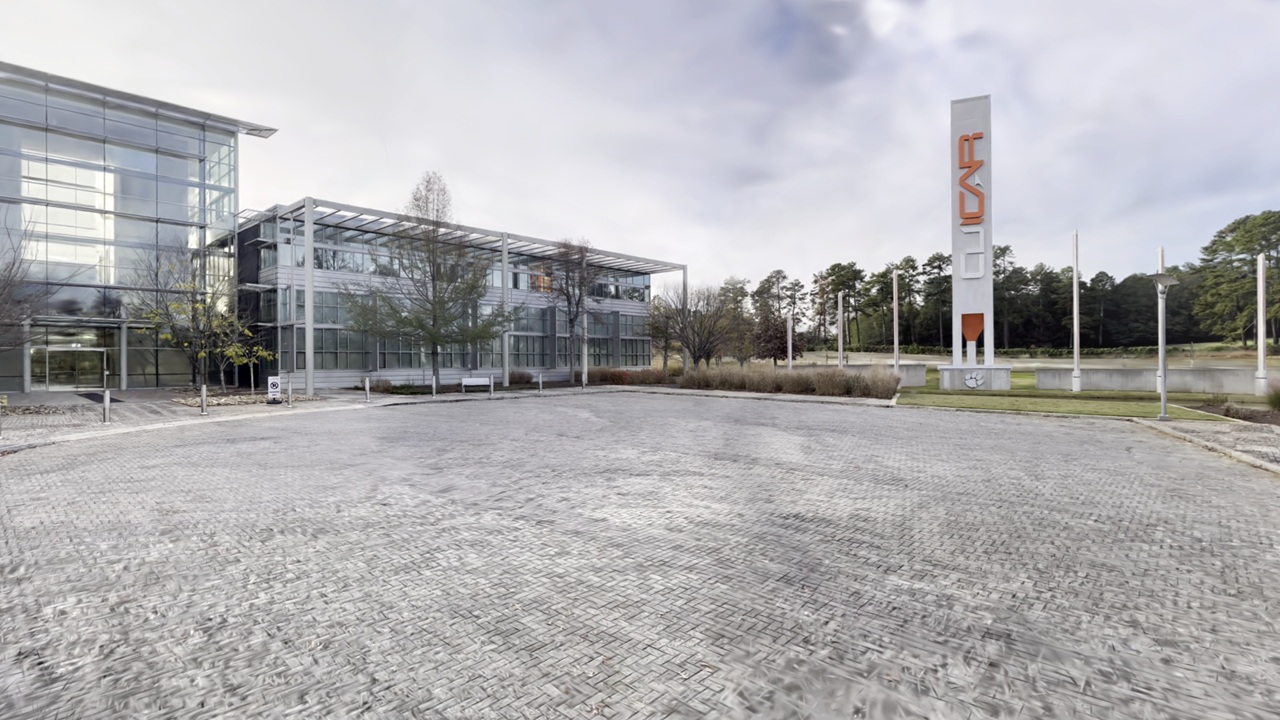}
         \caption{}
         \label{fig3a}
     \end{subfigure}
     \begin{subfigure}[b]{0.329\linewidth}
         \centering
         \includegraphics[width=\linewidth]{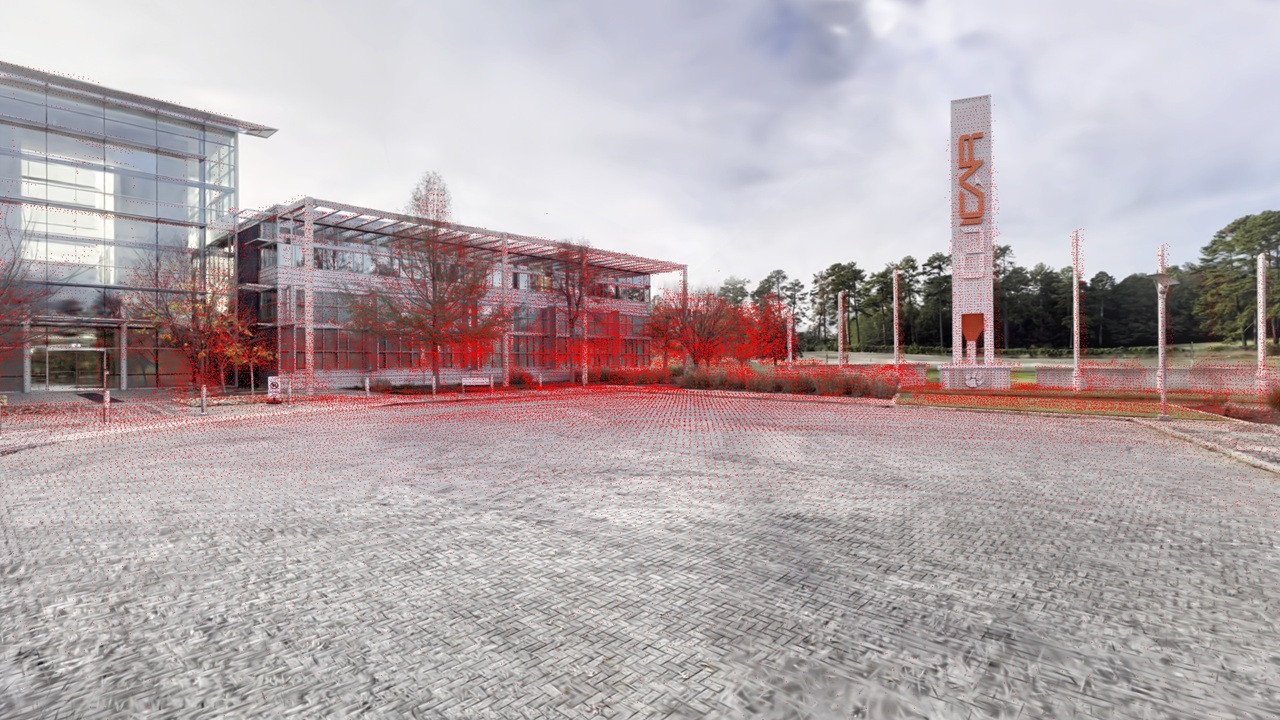}
         \caption{}
         \label{fig3b}
     \end{subfigure}
     \begin{subfigure}[b]{0.329\linewidth}
         \centering
         \includegraphics[width=\linewidth]{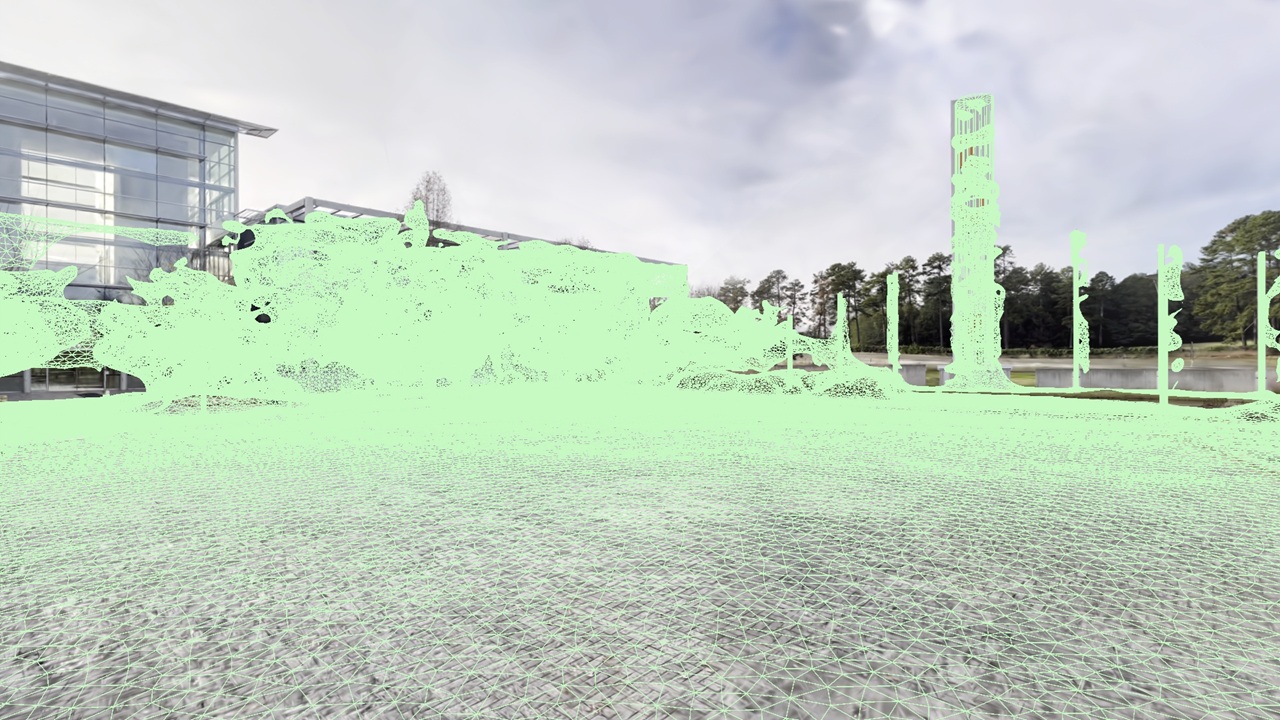}
         \caption{}
         \label{fig3c}
     \end{subfigure}
     \begin{subfigure}[b]{0.329\linewidth}
         \centering
         \includegraphics[width=\linewidth]{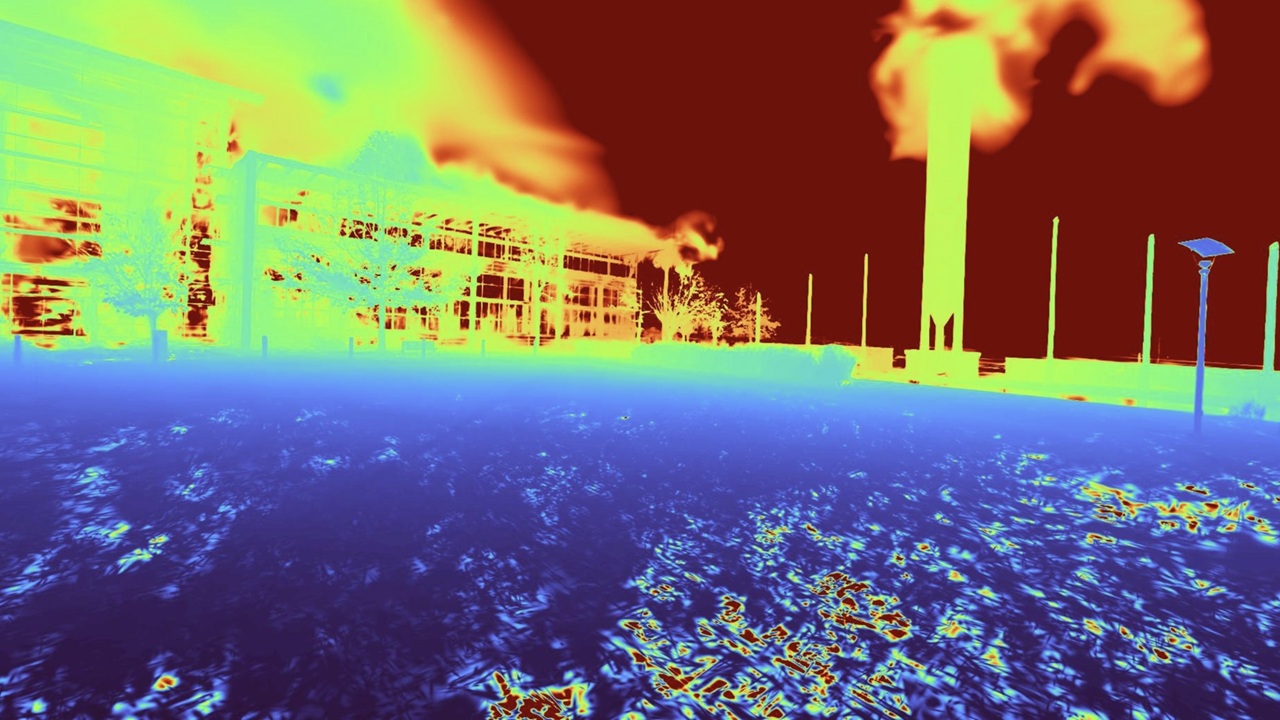}
         \caption{}
         \label{fig3d}
     \end{subfigure}
     \begin{subfigure}[b]{0.329\linewidth}
         \centering
         \includegraphics[width=\linewidth]{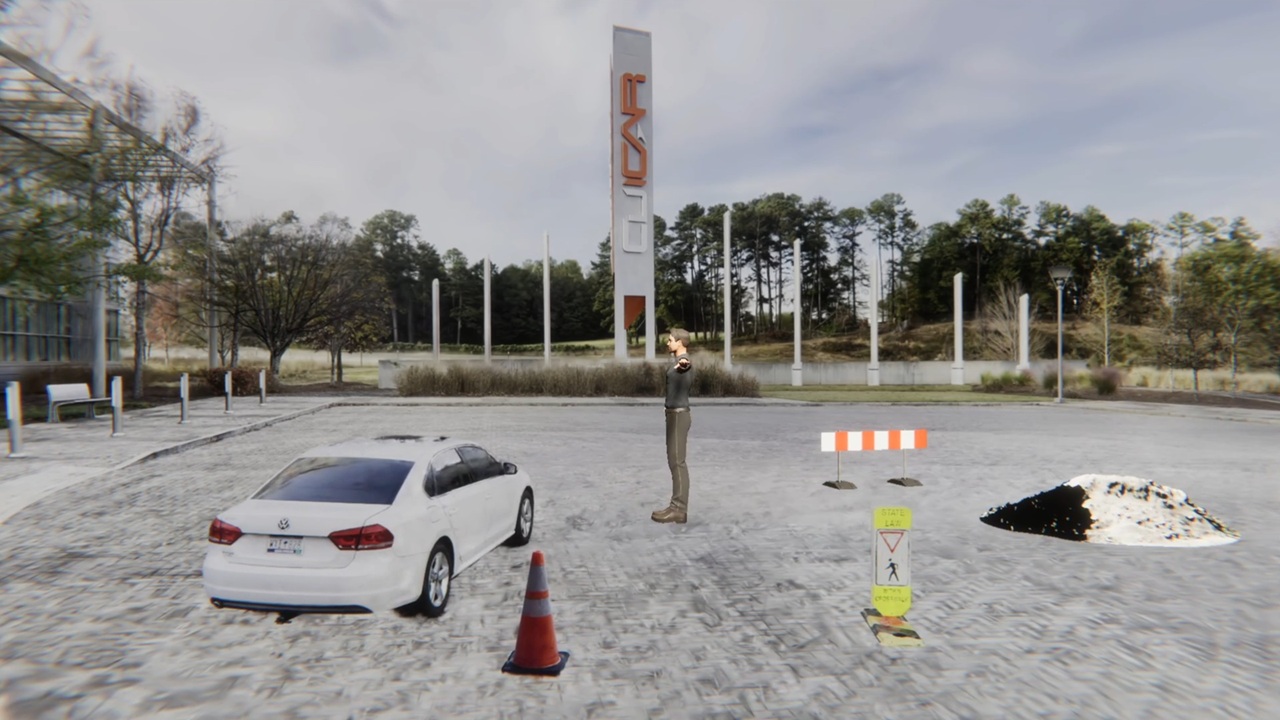}
         \caption{}
         \label{fig3e}
     \end{subfigure}
     \begin{subfigure}[b]{0.329\linewidth}
         \centering
         \includegraphics[width=\linewidth]{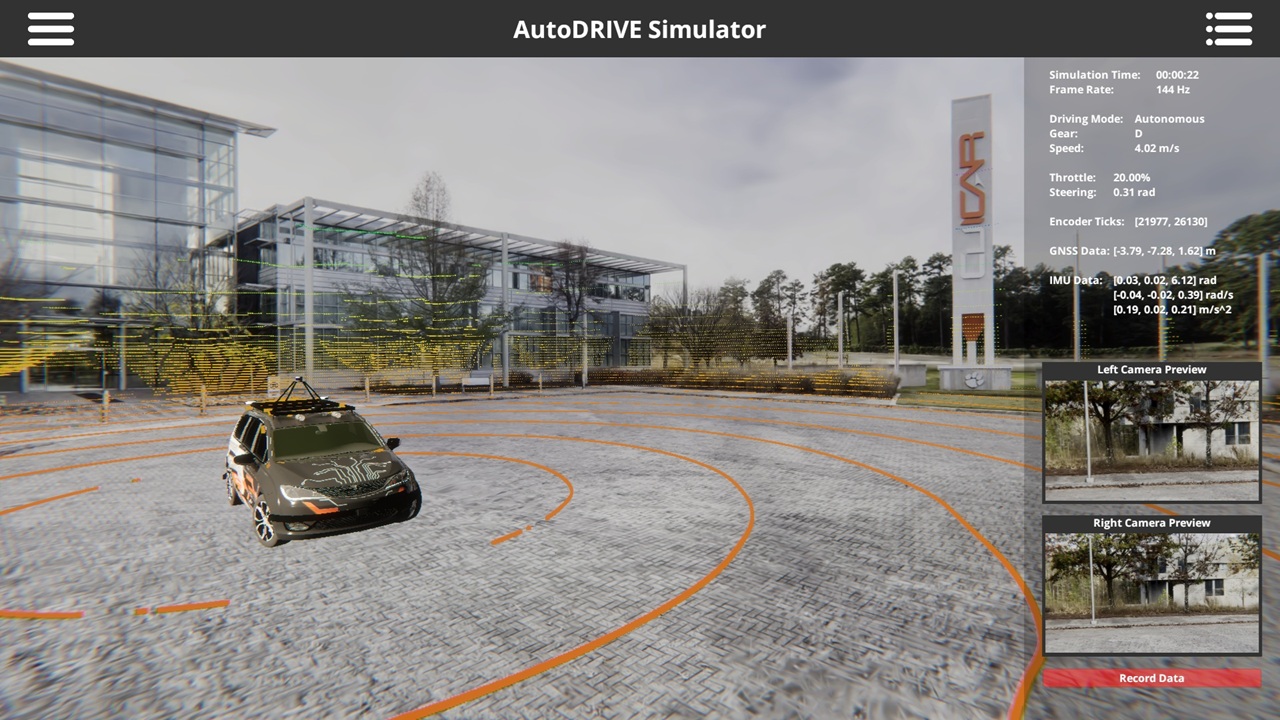}
         \caption{}
         \label{fig3f}
     \end{subfigure}
     \caption{High-fidelity reconstruction: Digital twin of the CU-ICAR campus showing (a) photorealistic rendering via 3DGS; (b) geometric accuracy w.r.t. 3D LIDAR point cloud data; (c) dynamical interaction via PSR; (d) visualization of reconstructed depth channel; (e) co-existence of 3DGS assets like a passenger car, a traffic cone, and a pedestrian sign with 3DMM assets like a road barrier, cement rubble, and a pedestrian; (f) real-time autonomy-oriented simulation with sensor visualization.}
    \label{fig3}
\end{figure*}

Scene reconstruction methods vary in complexity, accuracy, and application. Each method has unique strengths, with trade-offs between quality, speed, and computational cost. Manual hand-crafting has been shown to yield good results in some cases, but it is severely time-consuming, labor-intensive, and requires significant artistic skill. Additionally, manual reconstruction often introduces inaccuracies that affect the environment representation and sensor simulation. Surface reconstruction methods \cite{PoissonSurfaceRecon2006} generate continuous, water-tight 3D surfaces from sparse point clouds with high precision, but can struggle with noisy or incomplete data, and generally do not capture the visual appearance. Photogrammetry techniques \cite{pan2024globalstructurefrommotionrevisited} efficiently synthesize 3D models via images captured from multiple viewpoints by analyzing the overlap and spatial relationships between the images, delivering close geometric and visual resemblance, though they require precise image capture and may struggle with the reconstruction of large-scale assets and scenes. Neural radiance fields (NeRFs) \cite{NeRF2021} leverage modern deep learning methods to implicitly learn a continuous volumetric scene function from a sparse set of input views, generating 3D photorealistic reconstructions by synthesizing views from different perspectives. This method excels at reconstructing complex geometries and textures but demands high computational resources and is significantly slower in terms of rendering. 3D Gaussian splatting (3DGS) \cite{kerbl20233dgaussiansplattingrealtime} offers a fast and memory-efficient solution for scene reconstruction by employing lightweight 3D Gaussian functions to represent points in space as splats, parametrized by pose, scale, color, and opacity, which collectively contribute to the appearance and geometry of the reconstruction. This method often outperforms traditional voxel-based methods in terms of speed and quality, particularly excelling at reconstructing large-scale scenes and assets, which are central to autonomous driving simulations. However, there is no \textit{``one size fits all''} solution to scene/asset reconstruction. Hence, in this work, we propose to couple splat-based representations (for enhanced photorealism) with mesh-based representations (for dynamic interaction) to leverage their strengths and complement their weaknesses.

\subsection{Scenario Reconfiguration}

Scenario reconfiguration methods must strike a balance between customization, automation, flexibility, and computational efficiency. Manual design provides complete control over the design process to create highly customized, specific scenarios, but it is time-consuming, labor-intensive, and requires expert skill. Data-driven methods \cite{AutomatedDrivingScenarions2015} leverage large datasets to automatically generate or modify scenarios, offering scalability and efficiency, but are limited by the quality and diversity of the data used, potentially resulting in generic or unrealistic outcomes. Adversarial generation techniques \cite{9355111, 9578745} use generative models and search algorithms to create or transform scenarios into safety-critical corner or edge cases, but often require significant computational power and are sensitive to the quality of the sampled data. Knowledge-based generation \cite{8500632} utilizes rules, heuristics, or domain-specific information to generate scenarios, ensuring consistency and relevance, but can be less flexible and may struggle with unexpected scenarios. All in all, most of the aforementioned methods are very systematic and, therefore, limited in terms of creative exploration of novel scenarios. Consequently, recent works \cite{zhao2025surveyapplicationlargelanguage} have begun exploring LLM-based end-to-end scenario generation, yet many lack real-world grounding to audit realistic bounding and logic. Thus, in our work, we propose a hierarchical agentic AI framework comprising a mix of LLM agent(s) and rule-based agent(s) to generate driving scenarios on demand through user-convenient natural language prompts. The proposed method fosters the creativity of generative AI (LLMs can come up with really interesting suggestions, sometimes even beyond the user's imagination) while enforcing rule-based checks and bounds to ensure the generation of diverse yet pragmatic scenarios.

\begin{table}[t]
\centering
\caption{\small Comparison with State-of-the-Art Simulators}
\label{tab1}
\resizebox{\columnwidth}{!}{%
\begin{tabular}{l|c|c|c|c|c|c}
\hline
\multicolumn{1}{c|}{\textbf{Simulator}} & \multicolumn{1}{c|}{\textbf{\begin{tabular}[c]{@{}c@{}}Photorealistic\\ Rendering\end{tabular}}} & \multicolumn{1}{c|}{\textbf{\begin{tabular}[c]{@{}c@{}}Accurate\\ Dynamics\end{tabular}}} & \multicolumn{1}{c|}{\textbf{\begin{tabular}[c]{@{}c@{}}Hybrid Scene\\ Composition\end{tabular}}} & \multicolumn{1}{c|}{\textbf{\begin{tabular}[c]{@{}c@{}}LLM/VLM\\ Interface\end{tabular}}} & \multicolumn{1}{c|}{\textbf{\begin{tabular}[c]{@{}c@{}}Real-Time\\ Performance\end{tabular}}} & \multicolumn{1}{c}{\textbf{\begin{tabular}[c]{@{}c@{}}Open\\ Source\end{tabular}}} \\ \hline
CarSim \cite{CarSim}           & {\color[HTML]{FF0000} \xmark}    & {\color[HTML]{00A100} \cmark}    & {\color[HTML]{FF0000} \xmark}    & {\color[HTML]{FF0000} \xmark}    & {\color[HTML]{FFBB00} \smark}    & {\color[HTML]{FF0000} \xmark}    \\
Chrono \cite{ProjectChrono2016}           & {\color[HTML]{FF0000} \xmark}    & {\color[HTML]{00A100} \cmark}    & {\color[HTML]{FF0000} \xmark}    & {\color[HTML]{FF0000} \xmark}    & {\color[HTML]{FF0000} \xmark}    & {\color[HTML]{00A100} \cmark}    \\
GeoSim \cite{GeoSim2021}          & {\color[HTML]{00A100} \cmark}    & {\color[HTML]{FF0000} \xmark}    & {\color[HTML]{FF0000} \xmark}    & {\color[HTML]{FF0000} \xmark}    & {\color[HTML]{FFBB00} \smark}    & {\color[HTML]{FF0000} \xmark}    \\
MARS \cite{MARS2024}        & {\color[HTML]{00A100} \cmark}    & {\color[HTML]{FF0000} \xmark}    & {\color[HTML]{FF0000} \xmark}    & {\color[HTML]{FF0000} \xmark}    & {\color[HTML]{FF0000} \xmark}    & {\color[HTML]{00A100} \cmark}    \\
CARLA \cite{CARLA}        & {\color[HTML]{FFBB00} \smark}    & {\color[HTML]{FFBB00} \smark}    & {\color[HTML]{FF0000} \xmark}    & {\color[HTML]{FF0000} \xmark}    & {\color[HTML]{00A100} \cmark}    & {\color[HTML]{00A100} \cmark}    \\
LGSVL \cite{LGSVLSimulator}        & {\color[HTML]{FFBB00} \smark}    & {\color[HTML]{FFBB00} \smark}    & {\color[HTML]{FF0000} \xmark}    & {\color[HTML]{FF0000} \xmark}    & {\color[HTML]{00A100} \cmark}    & {\color[HTML]{FFBB00} \smark}    \\
\textbf{AutoDRIVE (Ours)}    & {\color[HTML]{00A100} \cmark}    & {\color[HTML]{00A100} \cmark}    & {\color[HTML]{00A100} \cmark}    & {\color[HTML]{00A100} \cmark}    & {\color[HTML]{00A100} \cmark}    & {\color[HTML]{00A100} \cmark}    \\ \hline

\multicolumn{7}{l}{\color[HTML]{00A100} \cmark \color{black} $\!$ indicates complete fulfillment; \color[HTML]{FFBB00} \smark \color{black} $\!$ indicates conditional or partial fulfillment; and \color[HTML]{FF0000} \xmark \color{black} $\!$ indicates non-fulfillment.} \\
\end{tabular}%
}
\end{table}

\subsection{Autonomy-Oriented Simulation}

Simulation frameworks vary in terms of their focus and fidelity. In the context of autonomous driving, the foci can be broadly categorized into dynamics and graphics. On one end of this spectrum, certain frameworks (e.g., \cite{CarSim, ProjectChrono2016}) emphasize vehicle dynamics or terramechanics simulations, ranging from multi-body models to finite- or discrete-element methods. Such frameworks usually compromise on visual fidelity or real-time performance. On the other end of the spectrum, some frameworks (e.g., \cite{GeoSim2021, MARS2024}) only focus on photorealistic rendering, and claim to achieve high fidelity through real-world data replays or graphical animations within reconstructed scenes. Such methods may not be real-time and are sometimes run offline. Finally, there are some frameworks (e.g., \cite{CARLA, LGSVLSimulator}) that fall between these two extremes and try to strike a balance between dynamics and graphics. These methods generally use simplified dynamics representations coupled with physically-based rendering (PBR) to compose driving scenarios with hand-crafted digital assets. Another challenge is that such simulated scenarios (static scenes or dynamic assets) are often not representative of real-world entities, which raises questions about simulation fidelity. All in all, none of the existing simulation frameworks support truly photorealistic rendering, high-fidelity dynamical simulation, and real2sim grounding, while offering extremely intuitive and user-friendly scenario reconfiguration and orchestration (refer Table \ref{tab1}). Our research aims to address this gap by proposing an openly accessible, high-fidelity yet real-time framework for democratizing AI-driven digital twins in autonomous driving by fusing literal photorealism and dynamical simulation with LLM-guided scenario editing.


\section{Research Methodology}
\label{Section: Research Methodology}

\subsection{Scene Reconstruction}

The proposed scene/asset reconstruction pipeline (refer Fig. \hyperref[fig2]{\ref*{fig2}(a-b)}) can leverage camera and LIDAR data from hand-held or vehicle-mounted sensors. Here, camera data contributes to photorealism, LIDAR data ensures geometric precision, and their combination enables hybrid scene composition.

The camera data is first preprocessed to extract frames from continuous video sequences, while ensuring consistent feature matching. We observed that camera resolution, exposure changes, and video frame rate significantly impact feature matching by directly affecting image sharpness, brightness/contrast, and motion blur. The extracted frames are parsed through a structure-from-motion (SfM) pipeline to obtain camera intrinsics, undistorted images, reconstructed poses, and feature points. The feature points are used to initialize 3D Gaussians. The camera intrinsics and reconstructed poses are used to project the 3D Gaussians from those viewpoints and rasterize 2D images. These rasterized images are compared with the corresponding undistorted images to compute the photometric loss. The gradients flow back via a differentiable tile rasterizer to update the 3D Gaussians (pose, scale, color, opacity) by minimizing error between the rasterized output and ground truth images.

Particularly, 3D Gaussian Splatting (3DGS) \cite{kerbl20233dgaussiansplattingrealtime} represents scenes and assets using a finite set of \textit{``splats''}, denoted as ${\mathcal{G}} = \{ \bm{g} \}$, where the intensity distribution of every splat follows a 3-dimensional Gaussian function. Each individual Gaussian, $\bm{g}({o}, {\bm{\mu}}, {\mathbf{q}}, {\bm{s}}, {\bm{c}})$ is parametrized by opacity ${o} \in [0, 1]$, position ${\bm{\mu}} \in \mathbb{R}^3$ represented as Cartesian coordinates, orientation ${\mathbf{q}} \in \mathbb{R}^4$ represented as a quaternion, anisotropic scaling factors ${\bm{s}} \in \mathbb{R}^3_+$, and view-dependent color representation ${\bm{c}} \in \mathbb{R}^F$ expressed as coefficients of spherical harmonics.
To compute the color $C$ of a rasterized pixel, $\mathcal{N}$ Gaussians that intersect with the pixel are sorted based on their distance to the camera center (sorted by index $i \in \mathcal{N}$) and blended using alpha compositing as:
\begin{align}
\label{3DGS-Render}
C = \sum_{i \in \mathcal{N}} {{\bm{c}}}_{i}\alpha_{i} \prod_{j=1}^{i-1} (1 - \alpha_{j})
\end{align}
where $\alpha_i = {o}_i \cdot \exp\left(-\frac{1}{2} (\mathbf{p} - \bm{\mu}_i)^T \bm{\Sigma}_i^{-1} (\mathbf{p} - \bm{\mu}_i)\right)$ for point $\mathbf{p}$ with $\bm{\Sigma}_i$ representing the 2D projection covariance.
The spatial information is captured by applying an affine transformation ${\mathbf{T}} = ({\mathbf{R}}, {\mathbf{t}}) \in \mathbb{SE}(3)$ to all Gaussians in the set as:
\begin{align}
\label{3DGS-Pose}
{\mathbf{T}} \otimes {\mathcal{G}} = { ({o}, {\mathbf{R}} {\bm{\mu}} + {\mathbf{t}}, \mathrm{Rot}({\mathbf{R}}, {\mathbf{q}}), {\bm{s}}, {\bm{c}}) }
\end{align}
where $\mathrm{Rot}(\cdot)$ represents the rotation of a quaternion using rotation matrix.

The LIDAR data is first registered to generate a cohesive point cloud, which is then segmented to filter out any unwanted noise and stray artifacts. Next, the point normals are estimated and aligned using a triangulation surface model. This post-processed (aligned) point cloud data (PCD) is used to reconstruct a 3D mesh model (3DMM) of the scene/asset.

Particularly, Poisson Surface Reconstruction (PSR) \cite{PoissonSurfaceRecon2006} recovers a 3D surface from oriented point clouds by solving a spatial Poisson equation. It treats aligned normals as samples of a vector field $\overrightarrow{V}$, seeking an implicit indicator function $\widetilde{\chi}$ whose gradient best matches $\overrightarrow{V}$ (i.e., $\nabla \widetilde{\chi} \approx \overrightarrow{V}$). By taking the divergence of the vector field formed by the normals, the problem reduces to solving the Poisson equation:
\begin{align}
\label{PSR-Problem}
\Delta \widetilde{\chi} \approx \nabla \overrightarrow{V}
\end{align}
The solution yields an approximation of the indicator function $\widetilde{\chi}$ in a least-squares sense. The reconstructed surface is extracted as an isosurface of this indicator function. PSR ensures smooth, watertight surfaces and is particularly robust to noise and non-uniform sampling, making it well-suited for geometric reconstruction of real-world 3D scenes and assets.

\begin{figure*}[t]
     \centering
     \begin{subfigure}[b]{0.245\linewidth}
         \centering
         \includegraphics[width=\linewidth]{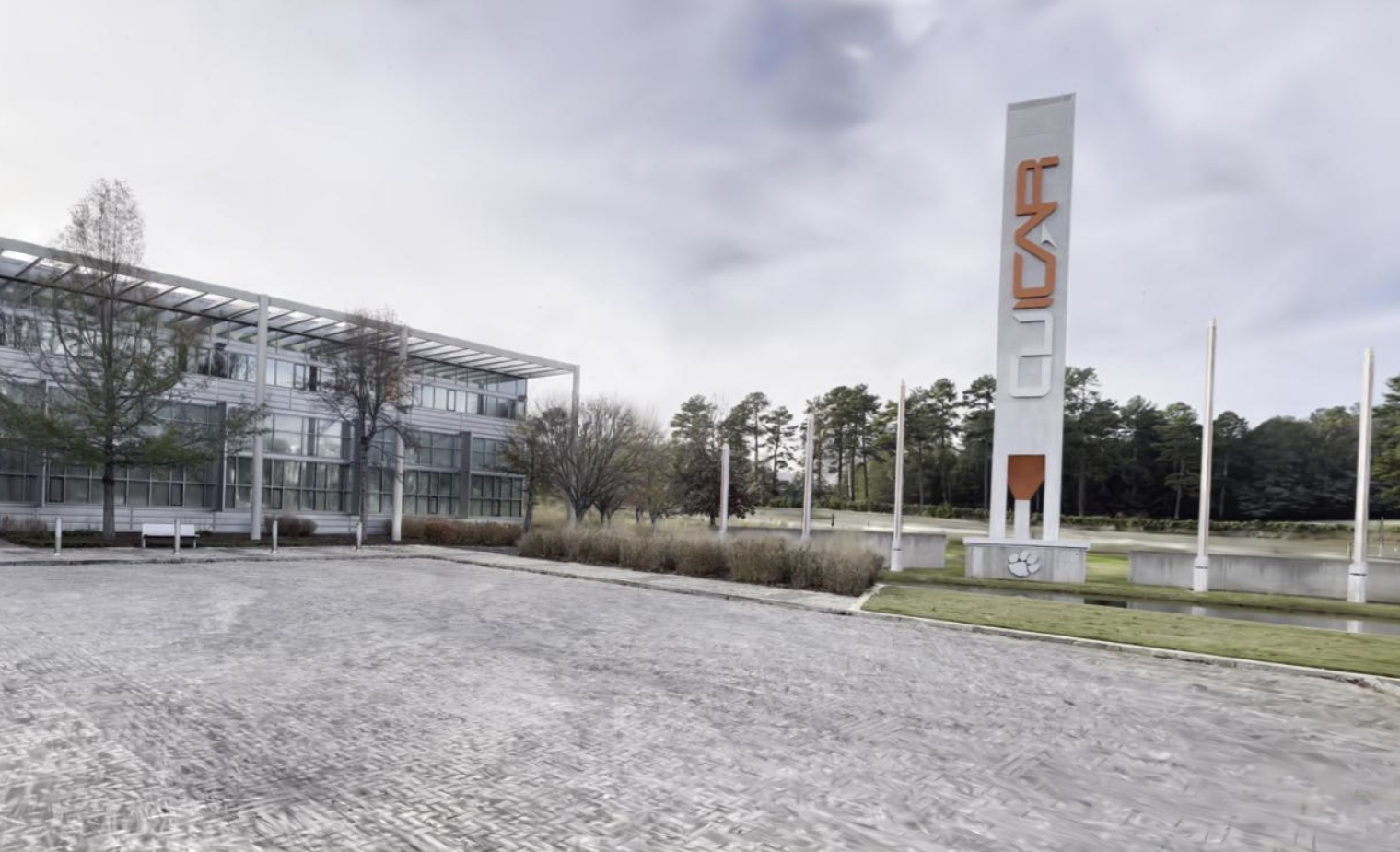}
         \caption{}
         \label{fig4a}
     \end{subfigure}
     \begin{subfigure}[b]{0.245\linewidth}
         \centering
         \includegraphics[width=\linewidth]{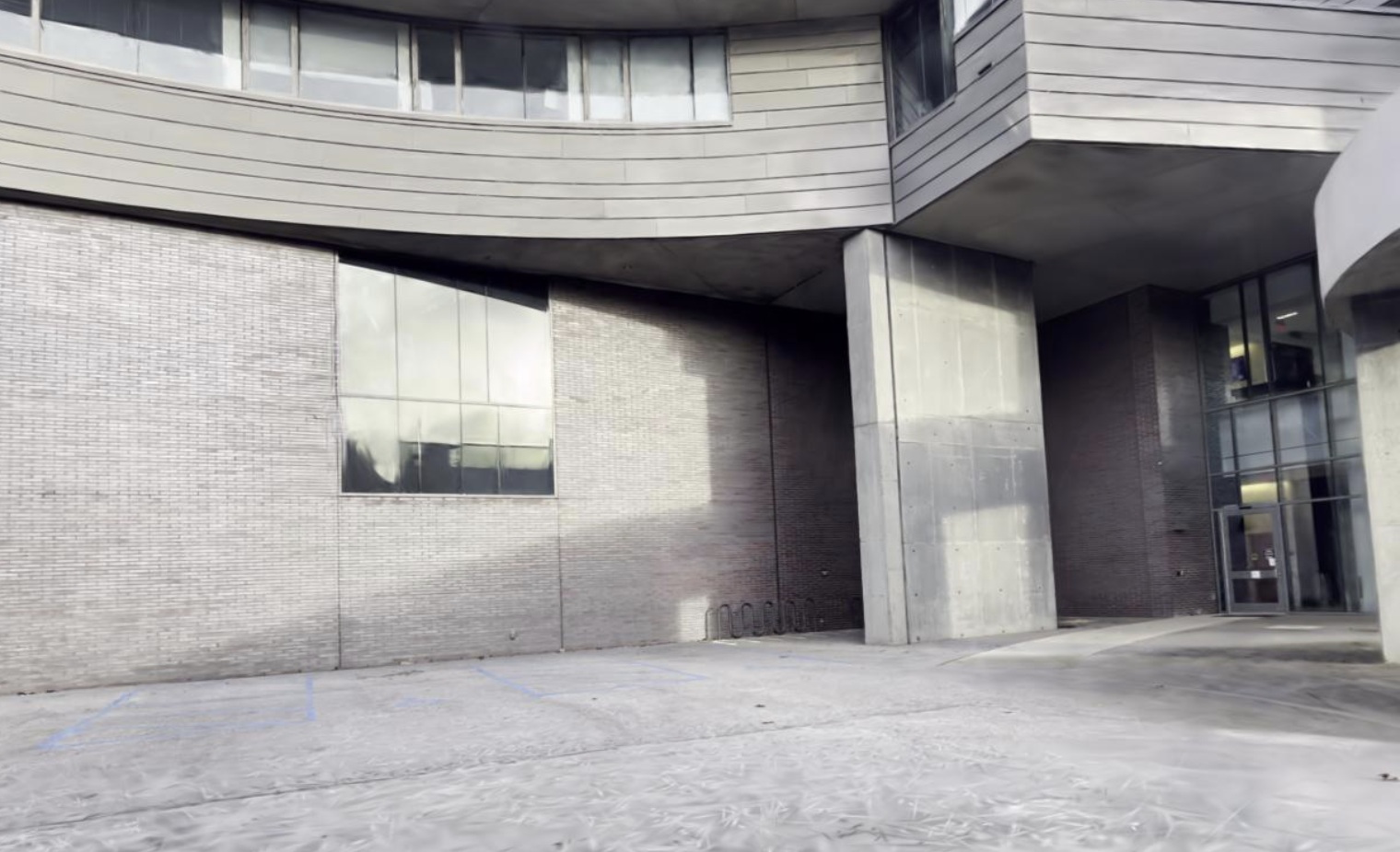}
         \caption{}
         \label{fig4b}
     \end{subfigure}
     \begin{subfigure}[b]{0.245\linewidth}
         \centering
         \includegraphics[width=\linewidth]{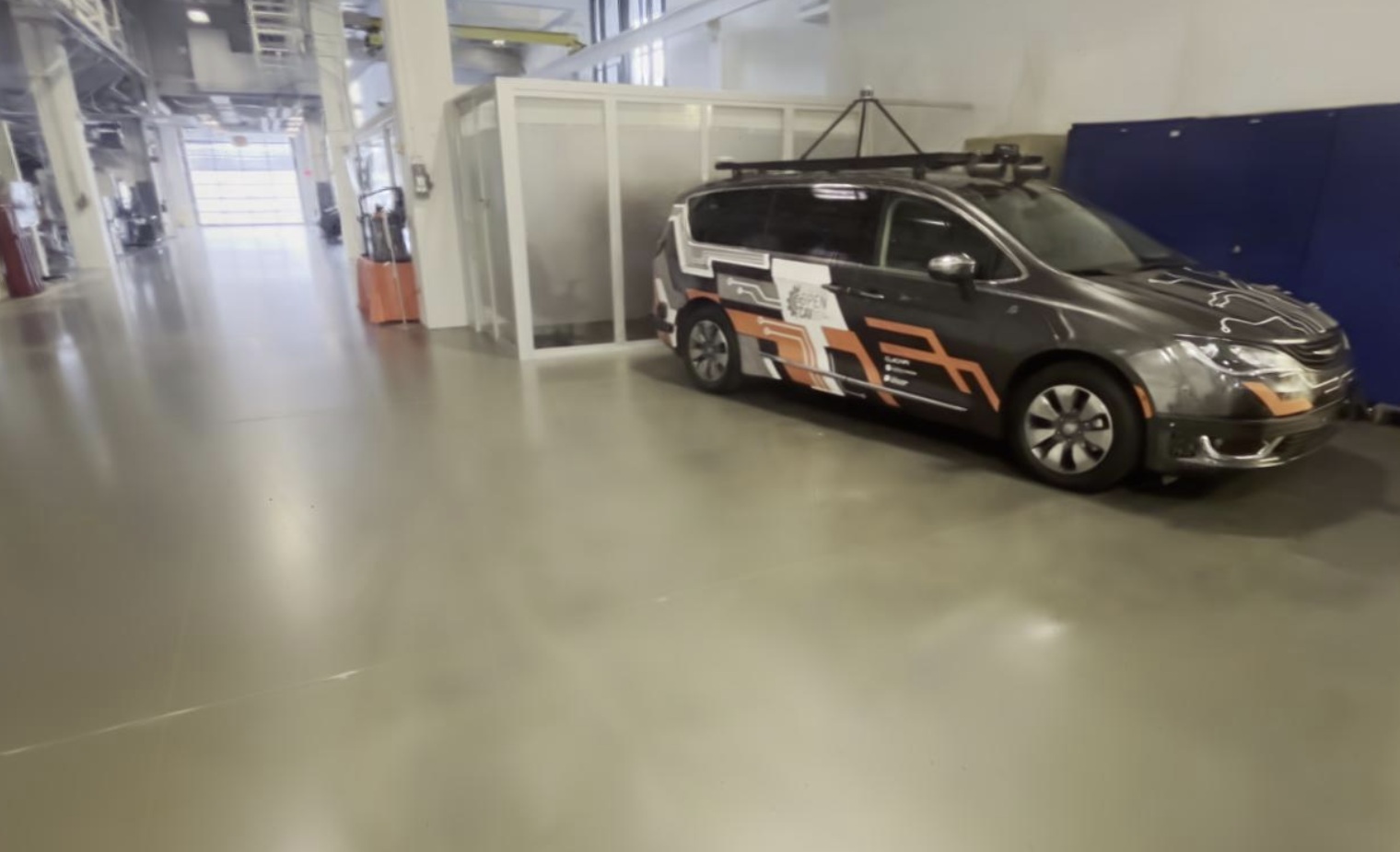}
         \caption{}
         \label{fig4c}
     \end{subfigure}
     \begin{subfigure}[b]{0.245\linewidth}
         \centering
         \includegraphics[width=\linewidth]{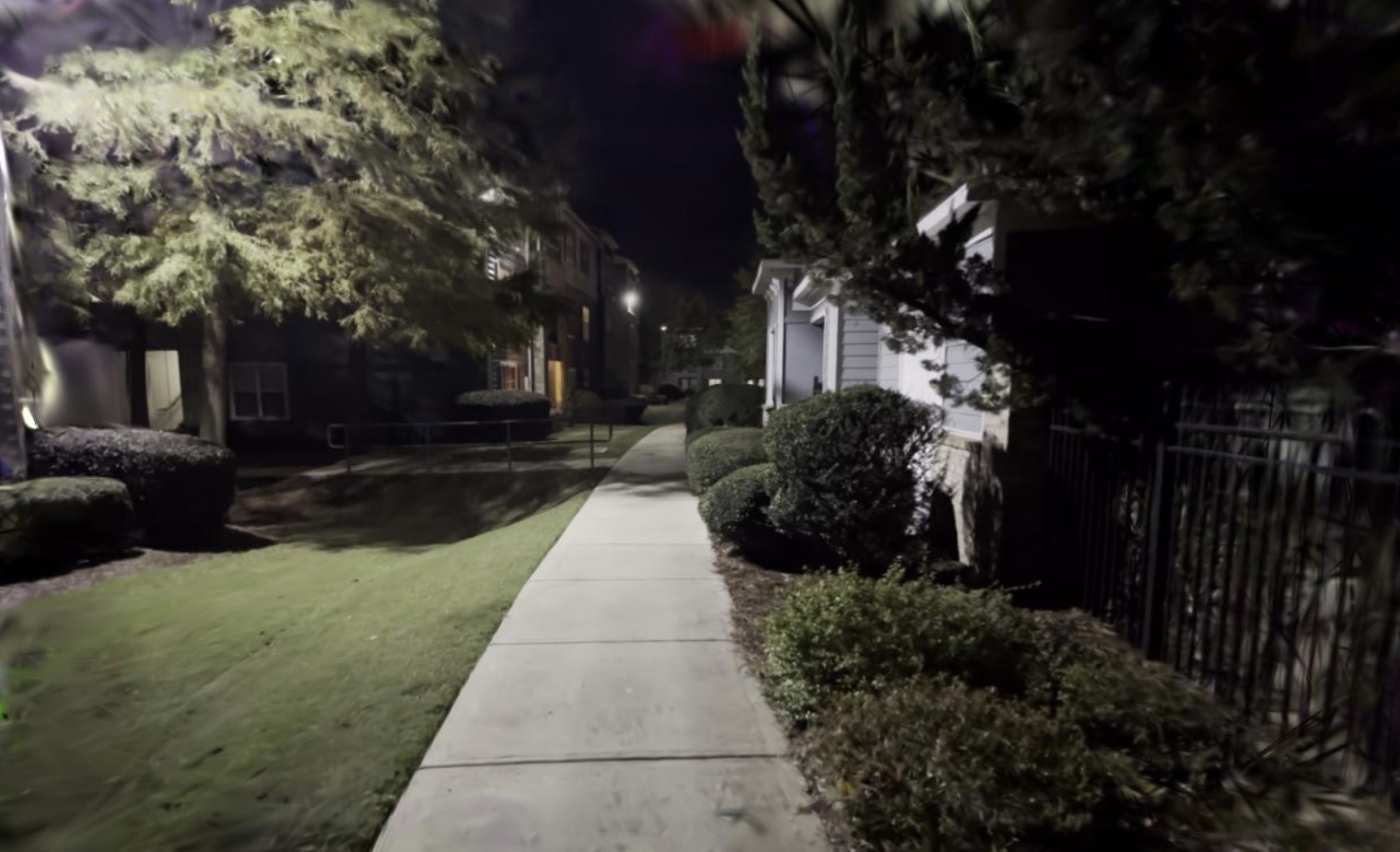}
         \caption{}
         \label{fig4d}
     \end{subfigure}
     \begin{subfigure}[b]{0.245\linewidth}
         \centering
         \includegraphics[width=\linewidth]{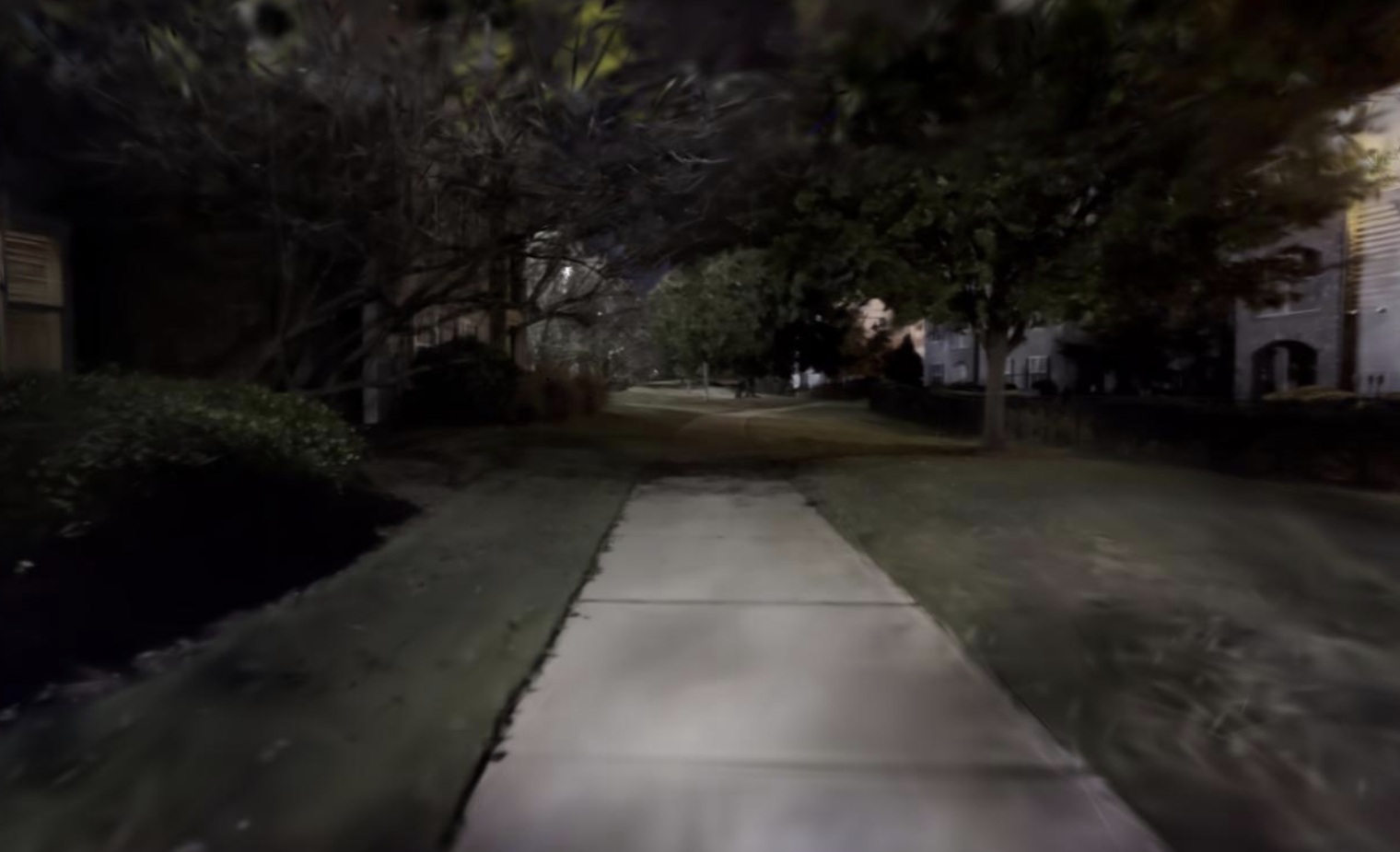}
         \caption{}
         \label{fig4e}
     \end{subfigure}
     \begin{subfigure}[b]{0.245\linewidth}
         \centering
         \includegraphics[width=\linewidth]{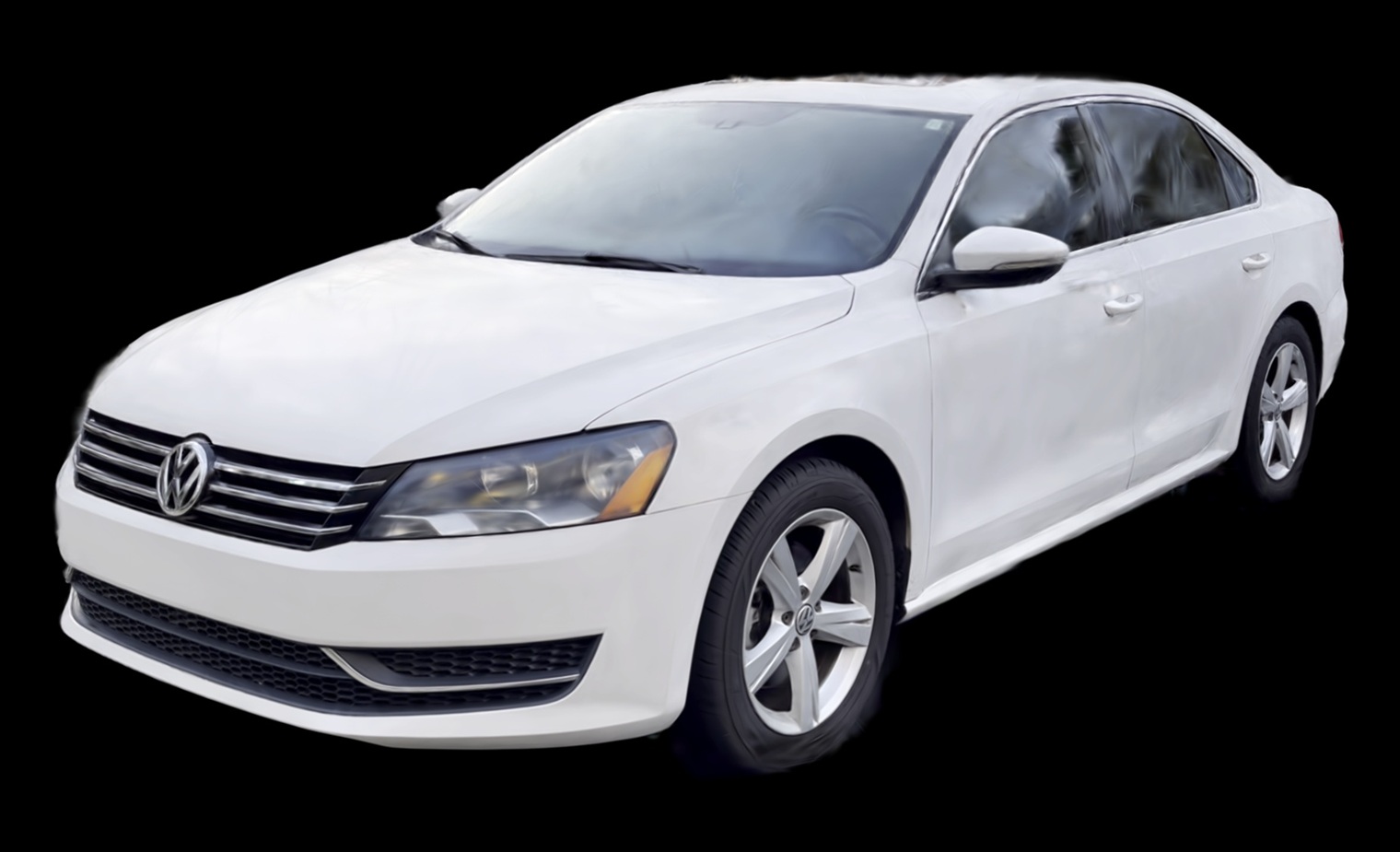}
         \caption{}
         \label{fig4f}
     \end{subfigure}
     \begin{subfigure}[b]{0.245\linewidth}
         \centering
         \includegraphics[width=\linewidth]{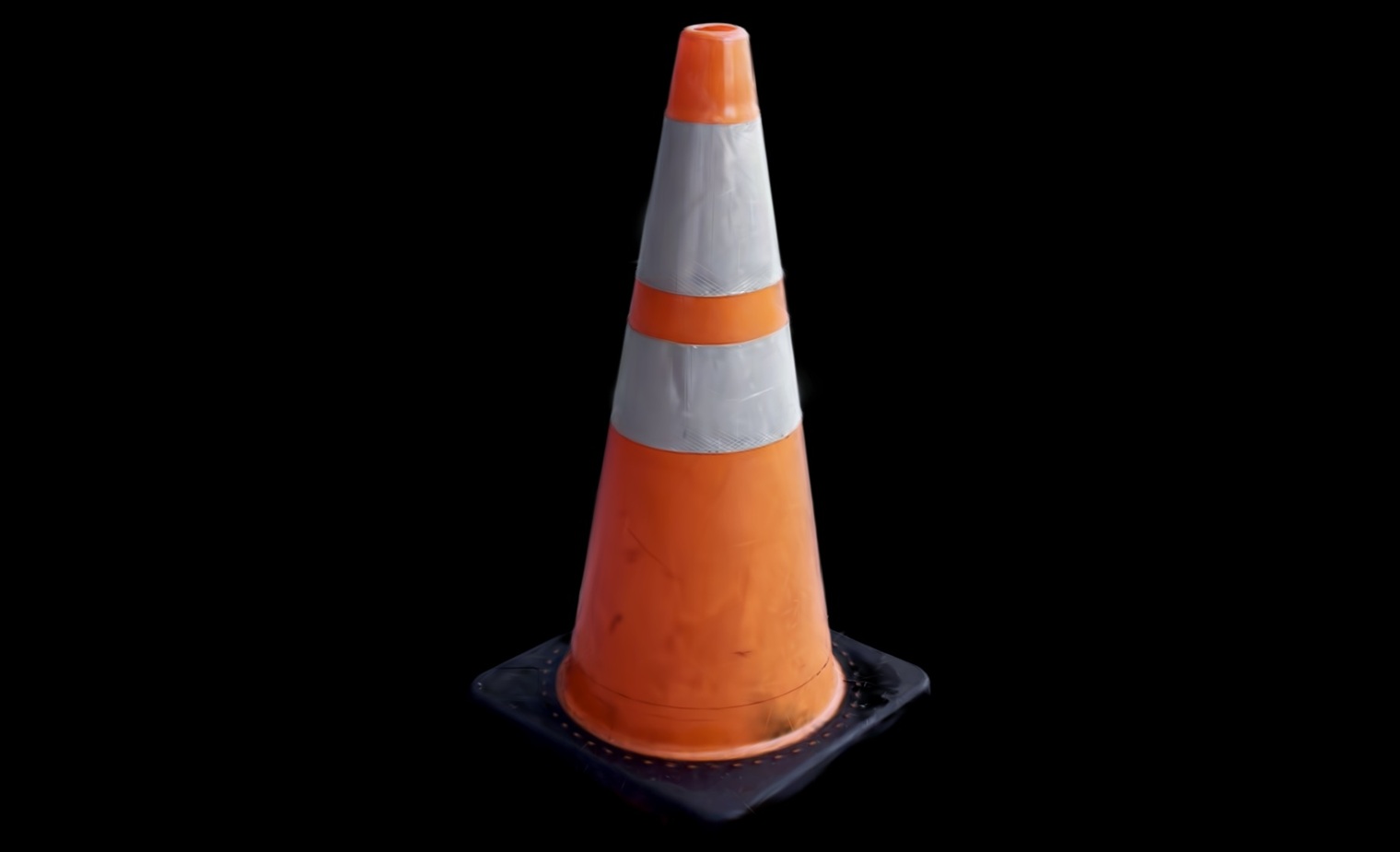}
         \caption{}
         \label{fig4g}
     \end{subfigure}
     \begin{subfigure}[b]{0.245\linewidth}
         \centering
         \includegraphics[width=\linewidth]{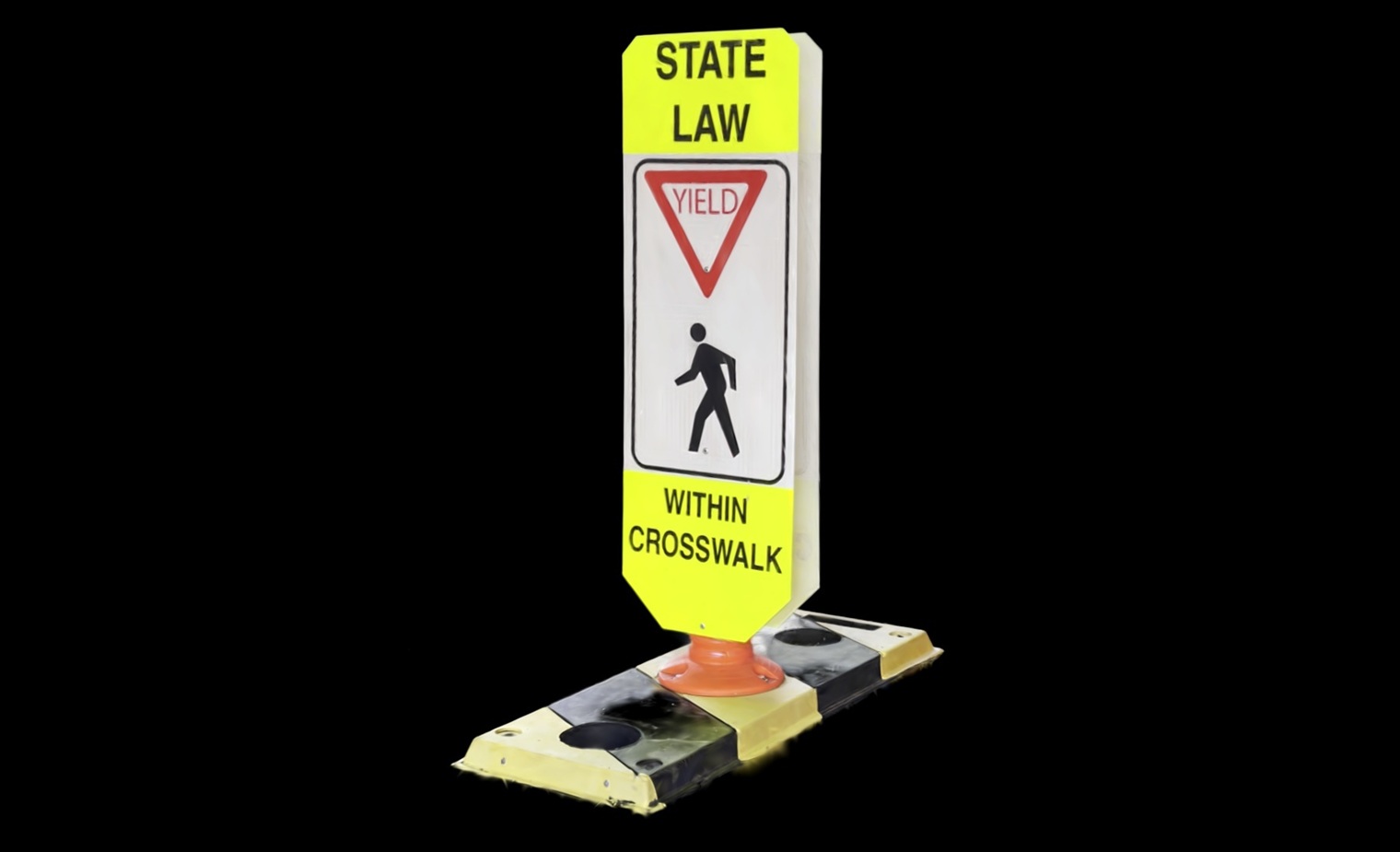}
         \caption{}
         \label{fig4h}
     \end{subfigure}
     \caption{Samples of reconstructed scenes (a) CU-ICAR, (b) CGEC, (c) AuE lab, (d) lit footpath, (e) unlit footpath; and assets (f) passenger car, (g) traffic cone, (h) pedestrian sign. Notice the shadows and reflections captured from real-world data.}
    \label{fig4}
\end{figure*}

The hybrid composition of 3DGS and PSR jointly achieves photorealism and physical validity in autonomous driving simulation. 3DGS provides a dense, view-dependent radiance representation with high-fidelity novel view synthesis, but its density-based, non-manifold structure lacks explicit surfaces, leading to unstable depth estimation. PSR, by contrast, reconstructs a globally consistent implicit surface from oriented PCD, yielding watertight meshes with well-defined normals suitable for collision detection and raycasting, but at the cost of limited appearance expressiveness. By decoupling appearance (3DGS) from physics (PSR) with geometrical grounding, the hybrid representation enforces cross-modal consistency across RGB-D channels, resulting in perceptually realistic and physically interactive scenes and assets.

The two reconstructed representations (3DGS and 3DMM) are fused to obtain \textit{``literally''} photorealistic (refer Fig. \hyperref[fig3]{\ref*{fig3}(a)}), geometrically accurate (refer Fig. \hyperref[fig3]{\ref*{fig3}(b)} and \hyperref[fig3]{\ref*{fig3}(d)}), and physically interactive (refer Fig. \hyperref[fig3]{\ref*{fig3}(c)}) 3D scenes/assets. This fusion is enabled by a hybrid rendering pipeline that comprises hierarchical passes for 3DGS rendering and physically based rendering (PBR). Consequently, these reconstructed scenes and assets can co-exist with other 3DGS or 3DMM assets and scenes (refer Fig. \hyperref[fig3]{\ref*{fig3}(e)}), enriching scenario generation capabilities. As an example, Fig \hyperref[fig3]{\ref*{fig3}(f)} demonstrates the reconstructed CU-ICAR campus (3DGS) within AutoDRIVE Simulator \cite{AutoDRIVESimulator2021}, simulating the autonomy-oriented digital twin of OpenCAV\footnote{\textbf{OpenCAV:} \url{https://sites.google.com/view/opencav}} \cite{DT-Across-Scales-2024} (3DMM) including vehicle dynamics, sensor physics, and environment interactions. Fig. \ref{fig4} depicts a few examples of reconstructed scenes and assets.

\subsection{Scenario Reconfiguration}

\begin{figure*}[t]
     \centering
     \includegraphics[width=\linewidth]{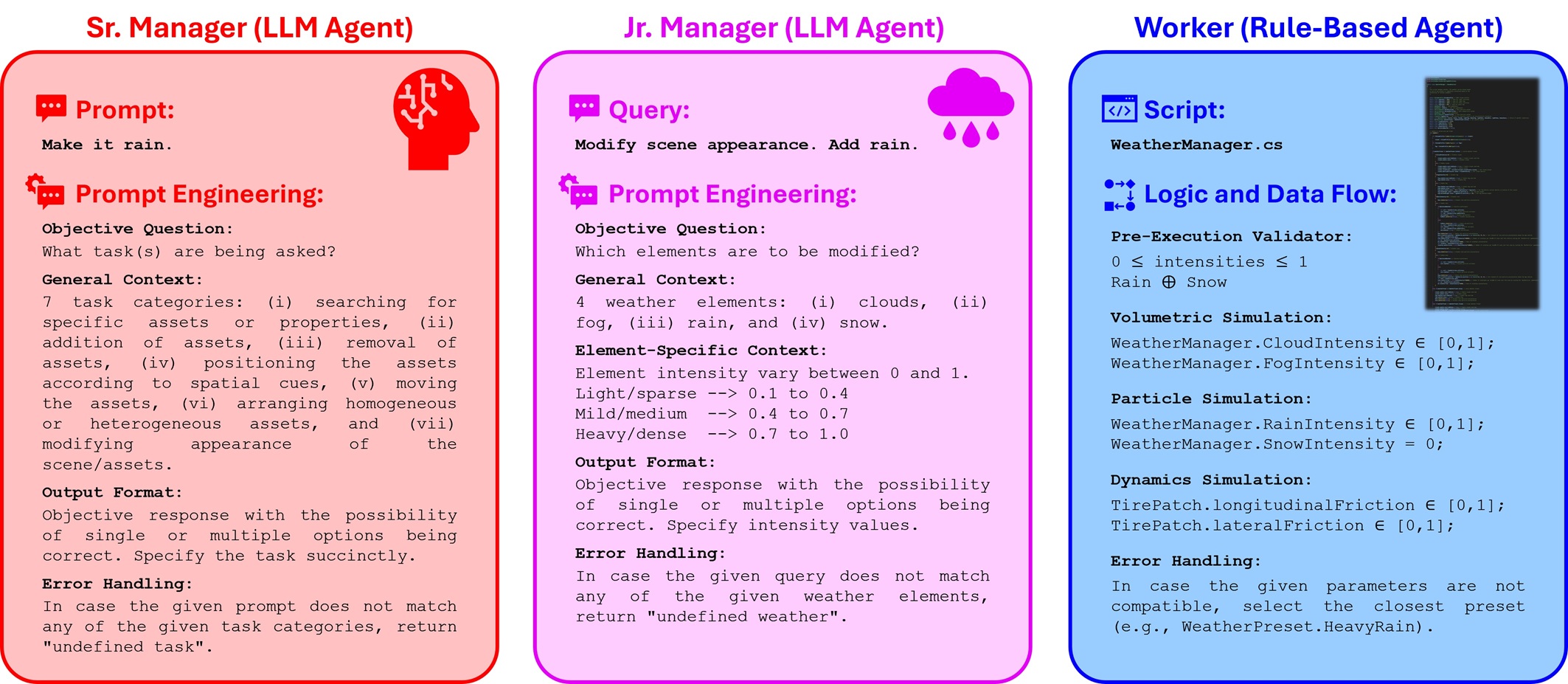}
     \caption{Worked example of scenario reconfiguration: (a) User prompt \textit{``make it rain''} being handled by level-1 LLM agent, (b) scenario design requirements being parsed by level-2 LLM agent, and (c) specialized tasks being implemented by rule-based agent to modify the existing scene.}
     \label{fig5}
\end{figure*}

The proposed scenario reconfiguration pipeline (refer Fig. \hyperref[fig2]{\ref*{fig2}(c)}) can leverage a hierarchical architecture comprising LLM agent(s) as well as rule-based agent(s). The level-1 LLM agent (a.k.a. ``Sr. Manager'') parses natural language inputs prompted by the human user(s) to determine the scenario design requirements based on specific keywords. An engineered prompt adds relevant context (based on the task/scene description, localization, and the asset library) as well as response format and error-handling instructions to the raw prompt. This ensures robust prompt handling and consistent response formatting for further levels of the hierarchy. The requirements are then passed on to level-2 LLM agents (a.k.a. ``Jr. Managers'') to determine the low-level tasks. Again, an engineered prompt adds relevant context, response format, and error-handling instructions to the raw query. These specialized tasks are then assigned to the appropriate rule-based agents (a.k.a. ``Workers'') to modify the existing scene by searching for, adding/removing/updating, or positioning/arranging the requested assets, or by incorporating specific properties/behaviors into them. The rule-based agents also implement pre-execution validation and error-handling mechanisms to avoid ill-posed scenario reconfiguration. This way, the proposed method fosters the creativity of generative AI, while enforcing rule-based constraints to ensure the generation of diverse yet pragmatic scenarios. It is worth noting that the agent hierarchy can be flexibly adjusted based on the expected complexity and granularity of scenario reconfiguration. Fig. \ref{fig5} presents a simple and intuitive worked example of scenario reconfiguration.

It is worth noting that scenario reconfiguration is an online, real-time process, meaning that the simulator continues to perform physics time-stepping, rendering, data transfer via application programming interfaces (APIs), accepting inputs, and providing feedback via the graphical user interface (GUI) or the human-machine interface (HMI), etc. Additionally, being an online process, the pipeline supports multi-stage commands while preserving context from the previous prompts. This is especially useful when sequentially editing the scenario while accepting or rejecting previous changes.

The proposed optional VLM-based visual enhancement pipeline leverages conditional latent diffusion models (e.g., SDXL). Particularly, the prompt engineering ensures that shading, lighting, reflections, shadows, and volumetric effects are blended throughout the scene without affecting the pose, scale, or other attributes of the scene/assets. It is also worth noting that this optional image enhancement can be performed online (e.g., to modify the rasterized output from the simulator, albeit at the cost of reduced frame rate) or offline (e.g., to batch-process synthetic datasets).


\section{Results and Discussion}
\label{Section: Results and Discussion}

\subsection{Scene Reconstruction}

The following metrics \cite{visapp24, Stocco2024} were used to assess the quality and performance of the scene/asset reconstructions:

\textbf{CCD/CMD:} Cloud-cloud distance (CCD) and cloud-mesh distance (CMD) quantify geometric reconstruction accuracy by measuring the point-wise (CCD) or point-to-surface (CMD) nearest-neighbor Euclidean distances (in meters) between a reconstructed point cloud and a reference point cloud (CCD), or reference mesh (CMD). Lower CCD/CMD values indicate better geometric alignment. CCD and CMD are sensitive to local geometric errors and can capture fine-scale surface discrepancies; however, they may be influenced by point density, noise, and alignment quality, and do not directly reflect the perceptual or semantic correctness.

\textbf{PSNR:} Peak signal-to-noise ratio (PSNR) quantifies the reconstruction fidelity by comparing the pixel-wise similarity between the rendered and the ground truth images. It focuses on pixel intensity differences and is derived from $L_2$ distance between the two images. Higher PSNR indicates better reconstruction quality. PSNR can measure even small pixel-level differences but may fail to capture perceptual distortions such as blurriness or loss of structural information.
    
\textbf{SSIM:} Structural similarity index measure (SSIM) evaluates the perceptual structural similarity between the rendered image and the ground truth by comparing luminance, contrast, and structure. It is designed to align with human visual perception. SSIM ranges from -1 to 1, indicating the degree of correlation between the pair of images (1 corresponds to perfect similarity). It captures perceptual quality better than PSNR by focusing on local image structures, which are critical for evaluating how well a method reconstructs edges, textures, and scene consistency.

\textbf{LPIPS:} Learned perceptual image patch similarity (LPIPS) is a deep-learning-based perceptual metric that compares high-level features extracted from a neural network (e.g., AlexNet in our case) rather than pixel-wise differences. It measures perceptual similarity between two images by computing the $L_2$ distance between their feature activations across multiple layers of a pre-trained neural network. LPIPS score ranges from 0 (perfect similarity) to 1 (completely different). It evaluates perceptual quality at a high level, capturing differences in texture, details, and global image structure.
    
\textbf{FPS:} Frames per second (FPS) is a performance measure to assess the rendering rate\footnote{All experiments were conducted using a single laptop PC with 12th Gen Intel Core i9-12900H 2.50 GHz CPU, NVIDIA GeForce RTX 3080 Ti GPU, and 32.0 GB RAM.}. It measures the number of frames rendered within a predefined time interval. FPS values can range from 0 (no update) to upwards of 30 (real-time), with $>$60 FPS being a good benchmark for autonomy-oriented simulations.

We first benchmark two different photorealistic reconstruction methods, viz. NeRF and 3DGS. Table \ref{tab2} presents the reconstruction results for the CU-ICAR scene (used as a benchmark). 3DGS outperforms NeRF by a large margin: not only are the quality metrics for 3DGS better than NeRF (PSNR +29.21\%, SSIM +90.24\%, LPIPS -64.29\%), but 3DGS is over 100 times faster than NeRF for the same scene. The choice of 3DGS is thus justified, and further results will assume 3DGS reconstruction, unless mentioned otherwise. Furthermore, as discussed earlier, we compare and propose to couple 3DGS-based photorealism (higher PSNR/SSIM and lower LPIPS) with PSR-based geometrical precision (lower CCD/CMD) and dynamic interactability (continuous surface representation) to complement each other.

\begin{figure*}[t]
     \centering
     \begin{subfigure}[b]{0.329\linewidth}
         \centering
         \includegraphics[width=\linewidth]{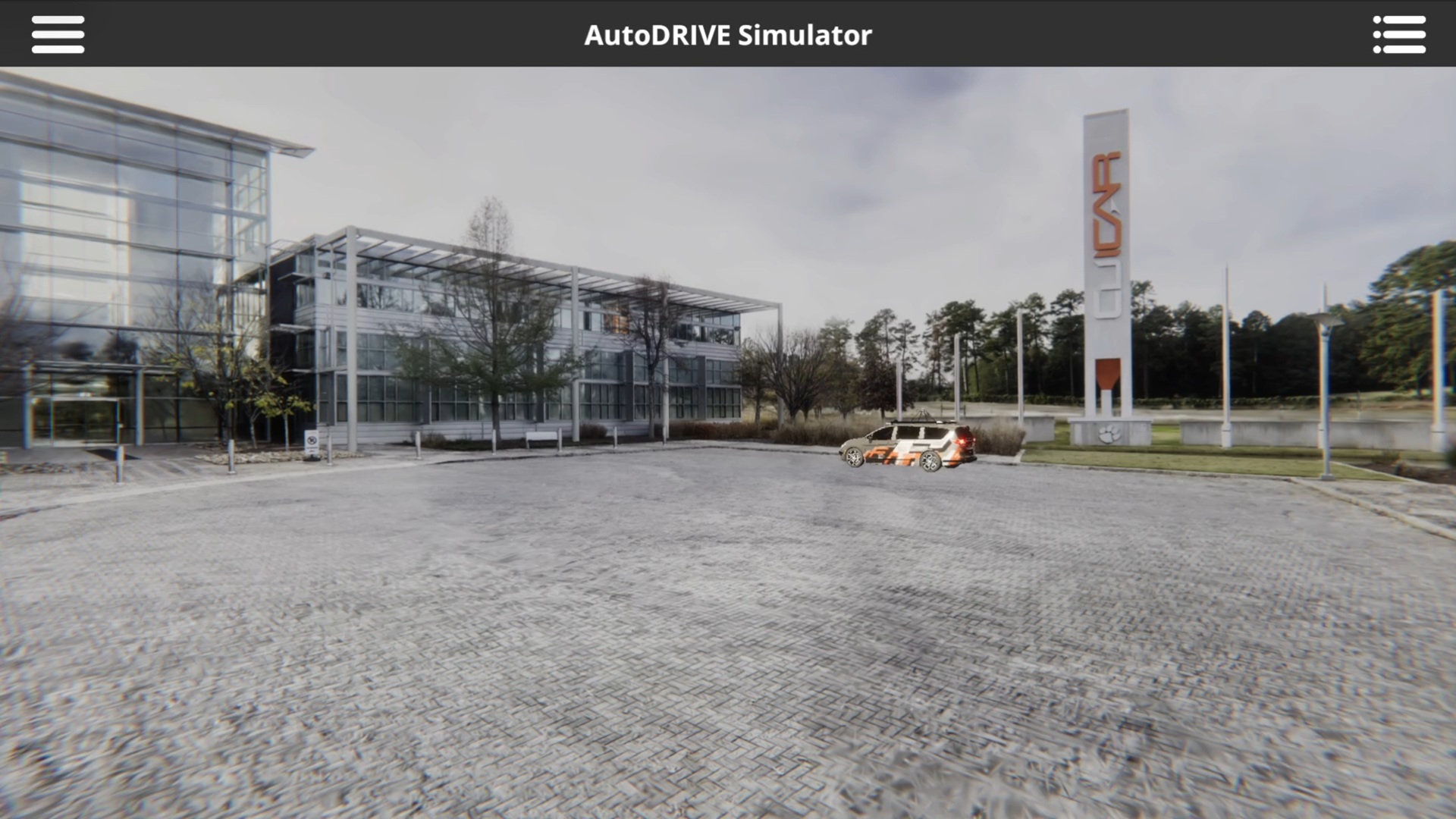}
         \caption{\textbf{Prompt:} It's daytime!}
         \label{fig6a}
     \end{subfigure}
     \begin{subfigure}[b]{0.329\linewidth}
         \centering
         \includegraphics[width=\linewidth]{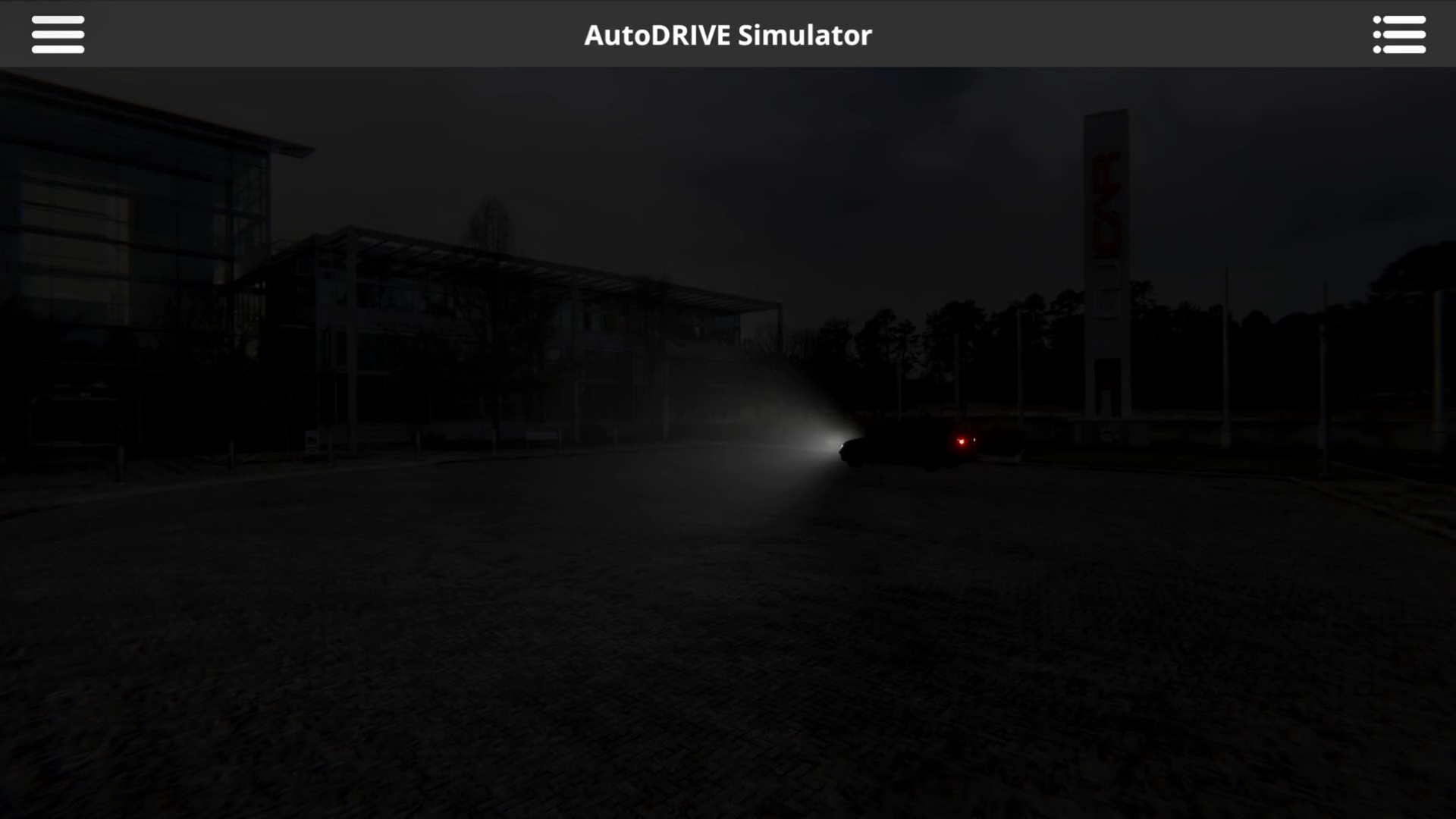}
         \caption{\textbf{Prompt:} Make it look like night time.}
         \label{fig6b}
     \end{subfigure}
     \begin{subfigure}[b]{0.329\linewidth}
         \centering
         \includegraphics[width=\linewidth]{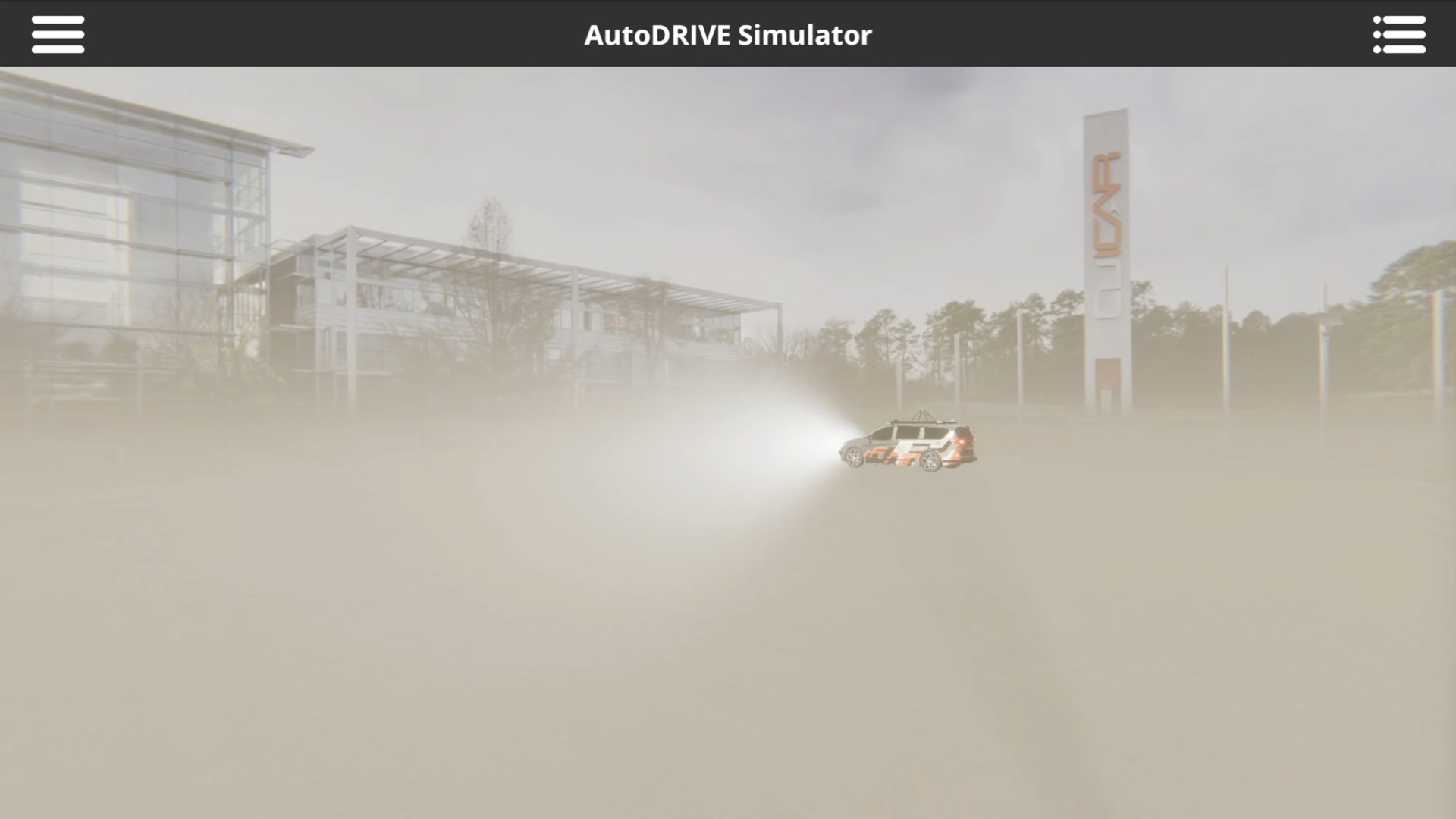}
         \caption{\textbf{Prompt:} Change weather to be foggy.}
         \label{fig6c}
     \end{subfigure}
     \begin{subfigure}[b]{0.329\linewidth}
         \centering
         \includegraphics[width=\linewidth]{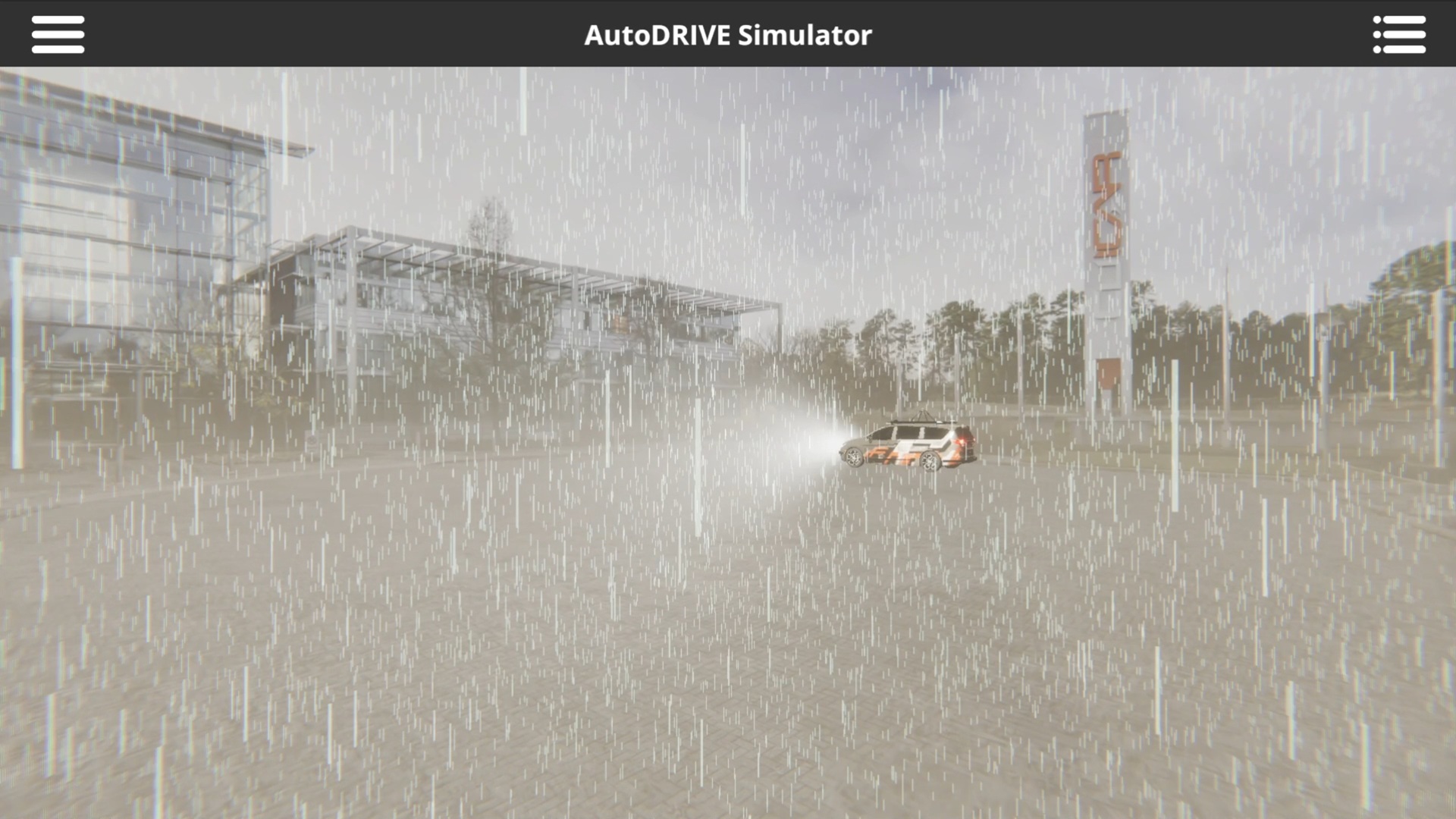}
         \caption{\textbf{Prompt:} Make it rain.}
         \label{fig6d}
     \end{subfigure}
     \begin{subfigure}[b]{0.329\linewidth}
         \centering
         \includegraphics[width=\linewidth]{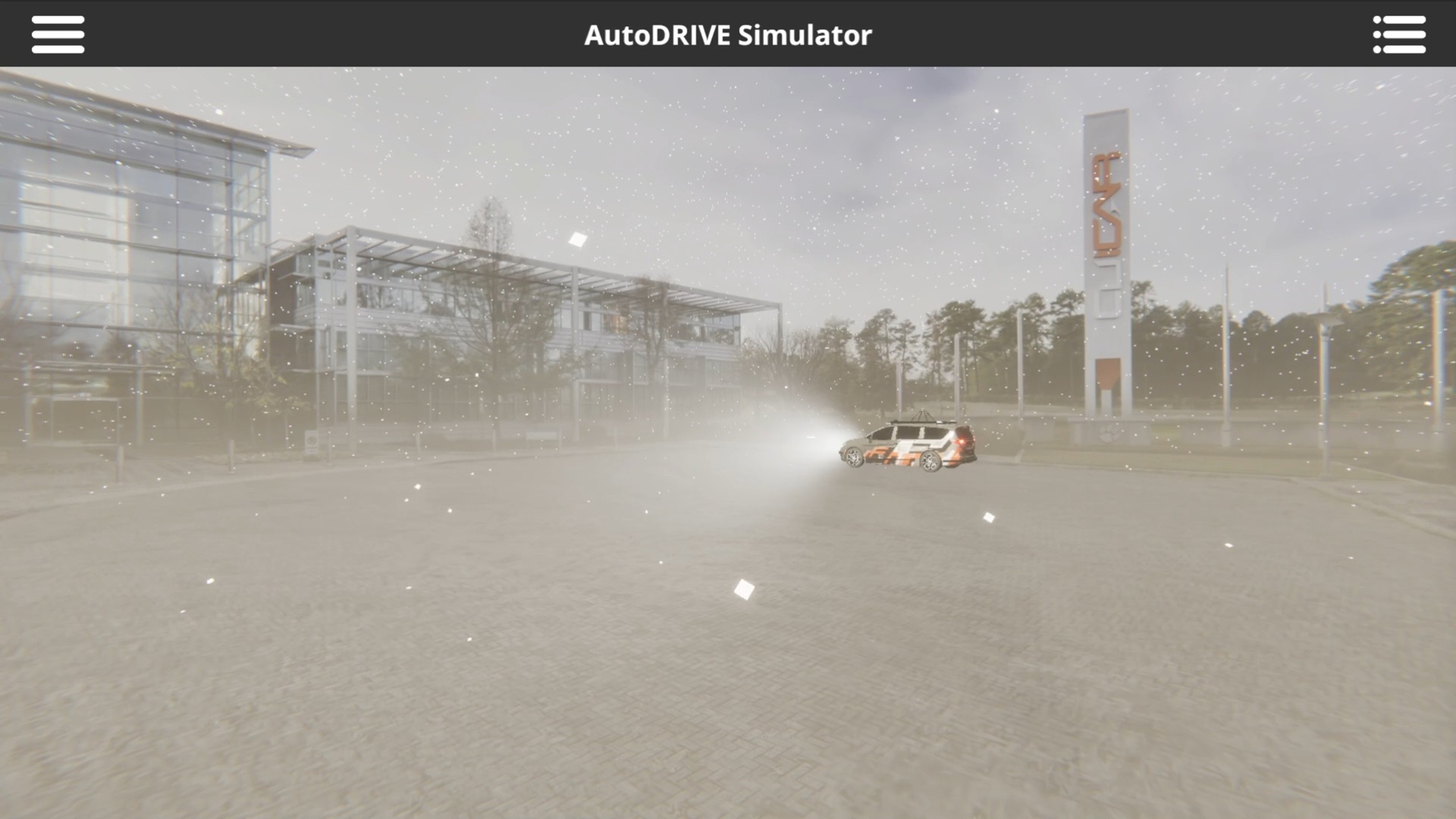}
         \caption{\textbf{Prompt:} Let it snow.}
         \label{fig6e}
     \end{subfigure}
     \begin{subfigure}[b]{0.329\linewidth}
         \centering
         \includegraphics[width=\linewidth]{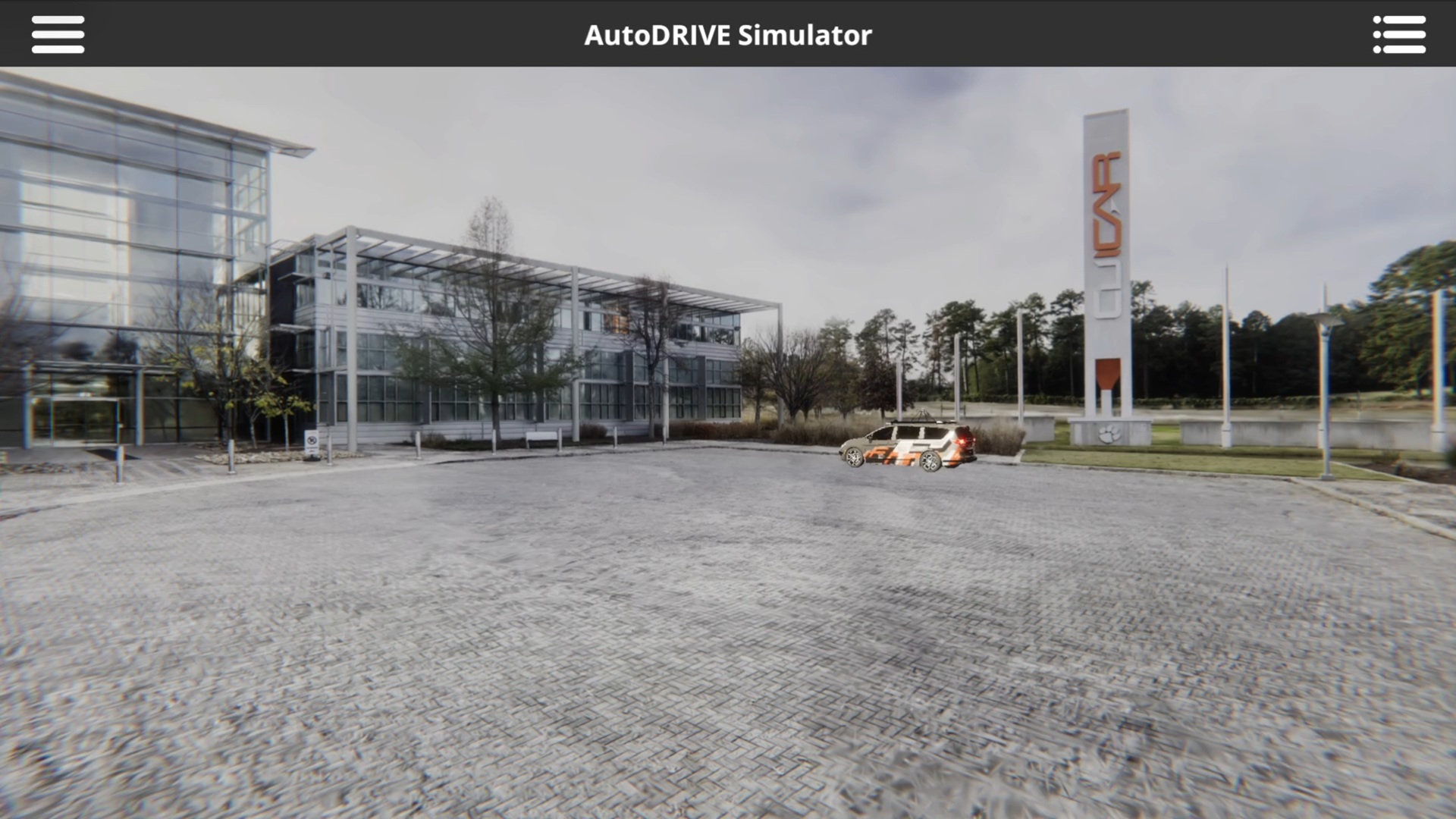}
         \caption{\textbf{Prompt:} It's a clear day.}
         \label{fig6f}
     \end{subfigure}
     \begin{subfigure}[b]{0.329\linewidth}
         \centering
         \includegraphics[width=\linewidth]{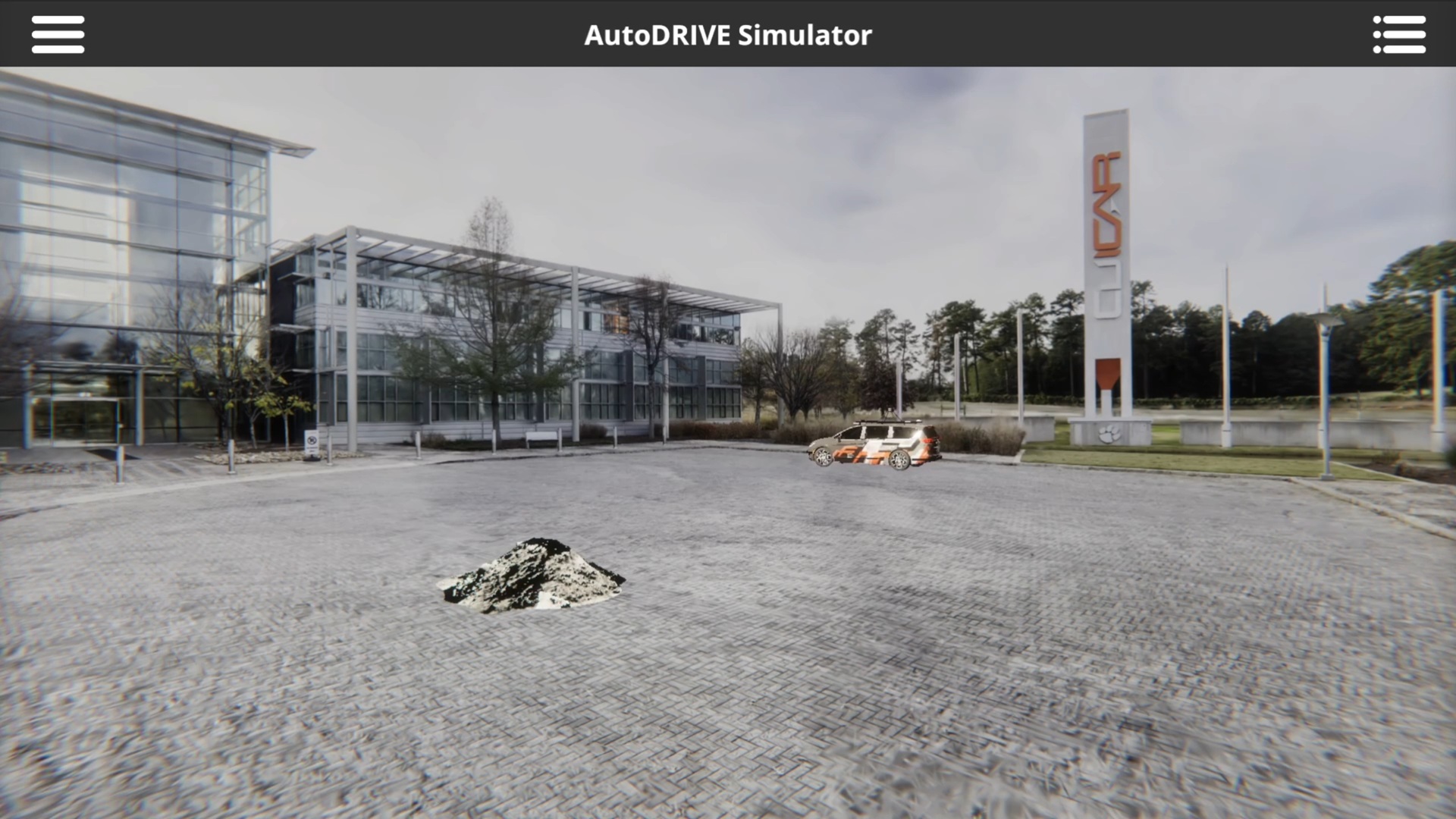}
         \caption{\textbf{Prompt 1:} Create cement rubble at the center of the scene.}
         \label{fig6g}
     \end{subfigure}
     \begin{subfigure}[b]{0.329\linewidth}
         \centering
         \includegraphics[width=\linewidth]{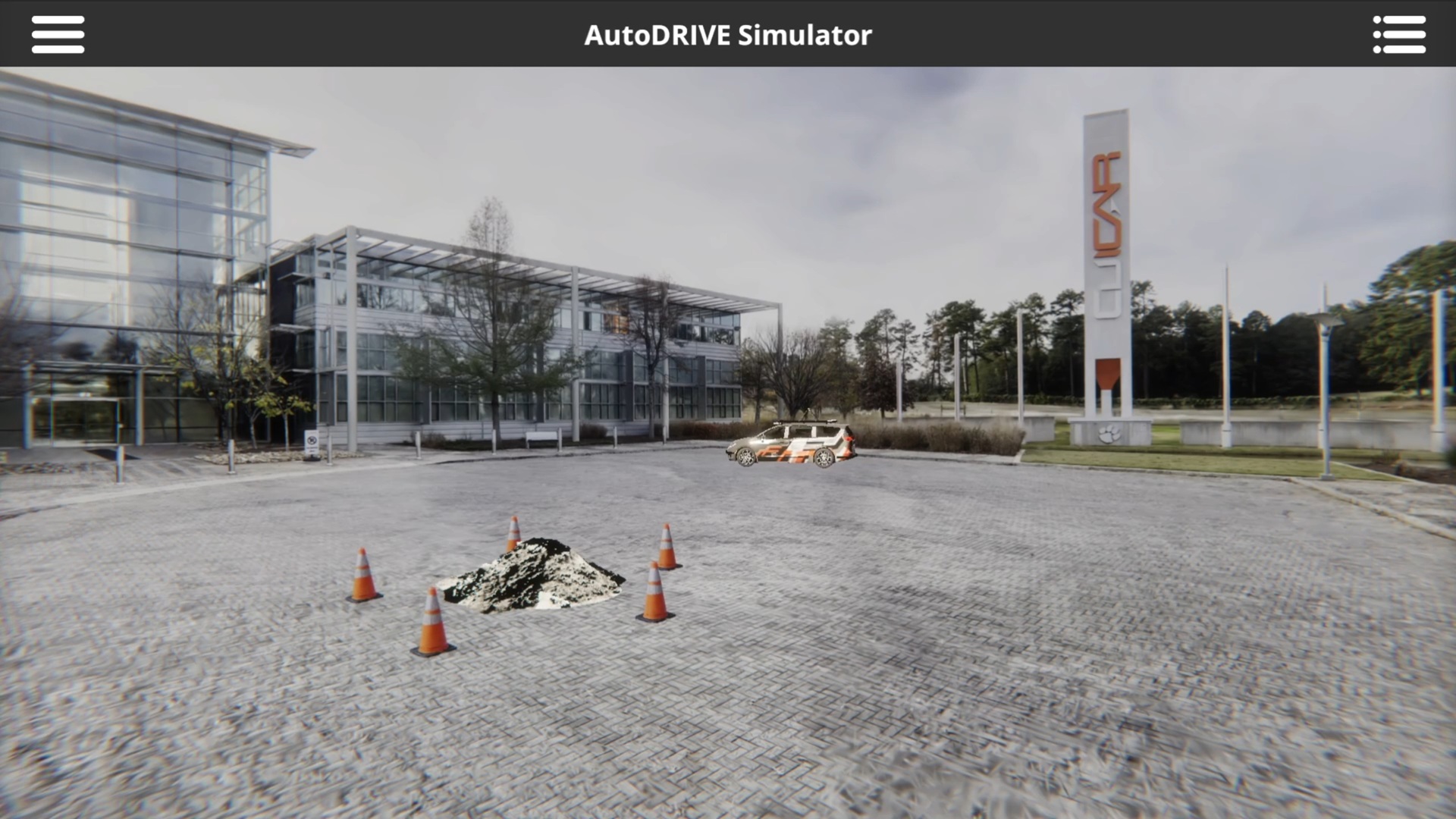}
         \caption{\textbf{Prompt 2:} Add traffic cones to mark the maintenance.}
         \label{fig6h}
     \end{subfigure}
     \begin{subfigure}[b]{0.329\linewidth}
         \centering
         \includegraphics[width=\linewidth]{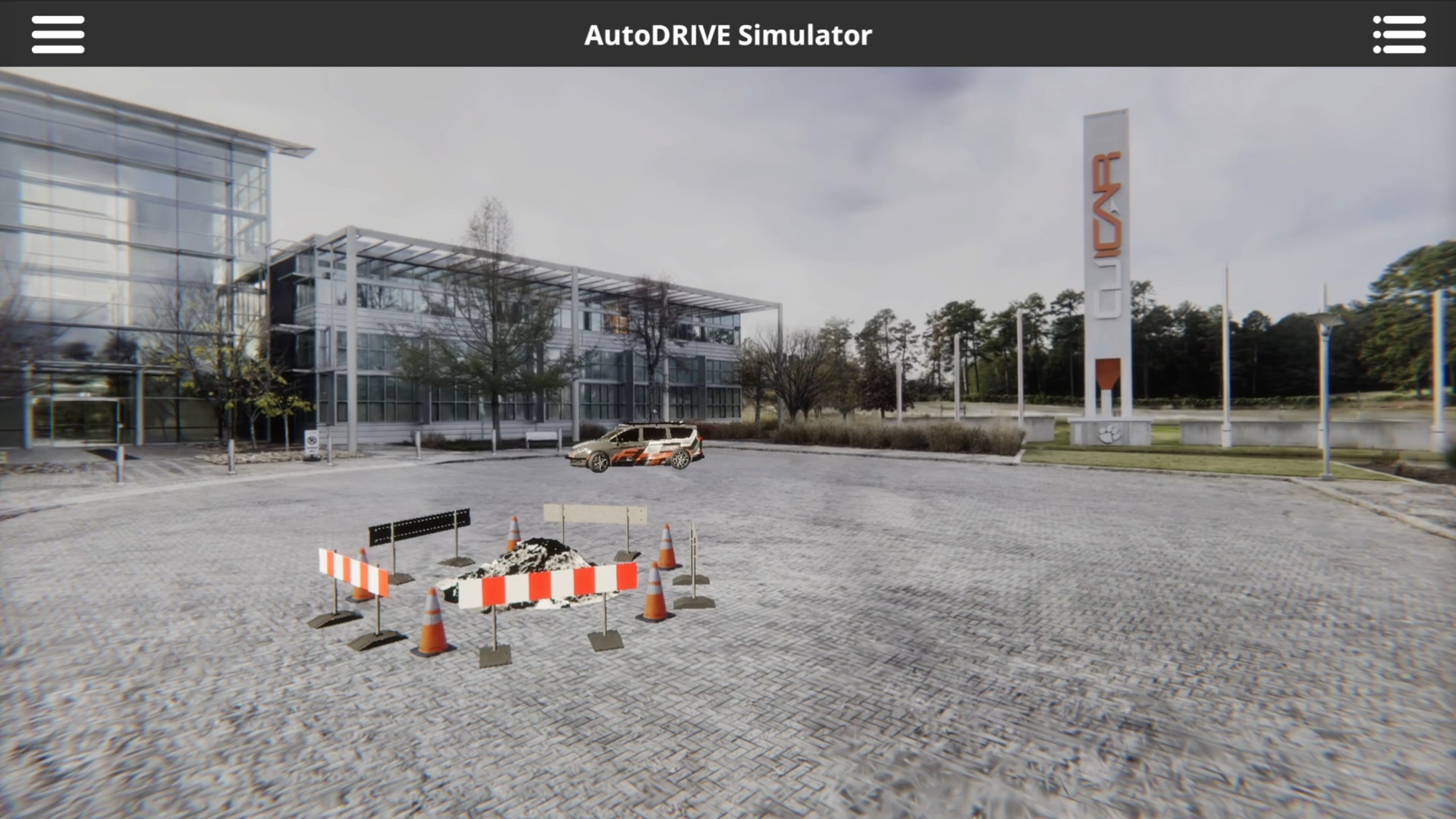}
         \caption{\textbf{Prompt 3:} Also add road barriers around there.}
         \label{fig6i}
     \end{subfigure}
     \begin{subfigure}[b]{0.329\linewidth}
         \centering
         \includegraphics[width=\linewidth]{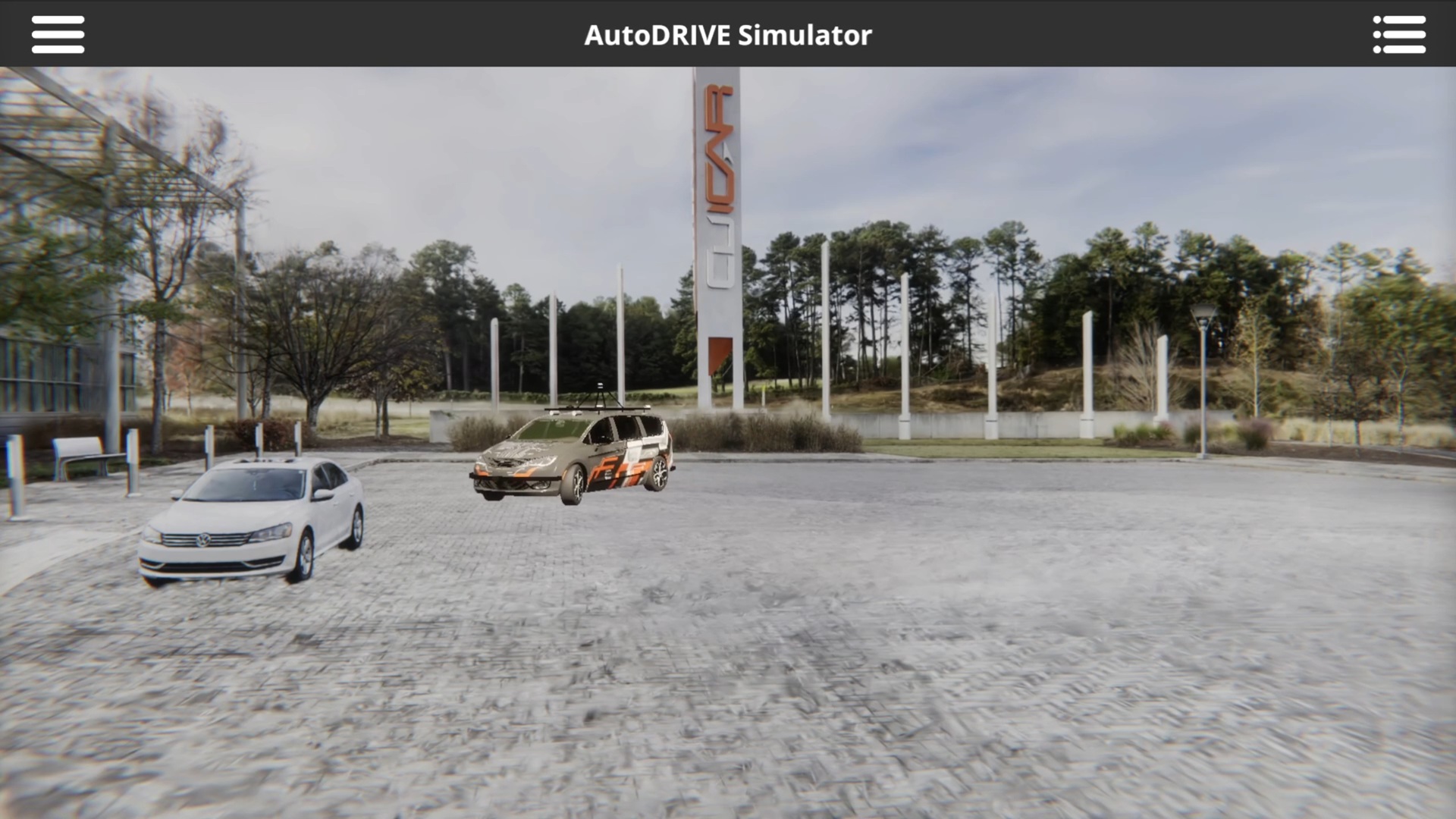}
         \caption{\textbf{Prompt:} Add a parked car at the JTEKT building entrance.}
         \label{fig6j}
     \end{subfigure}
     \begin{subfigure}[b]{0.329\linewidth}
         \centering
         \includegraphics[width=\linewidth]{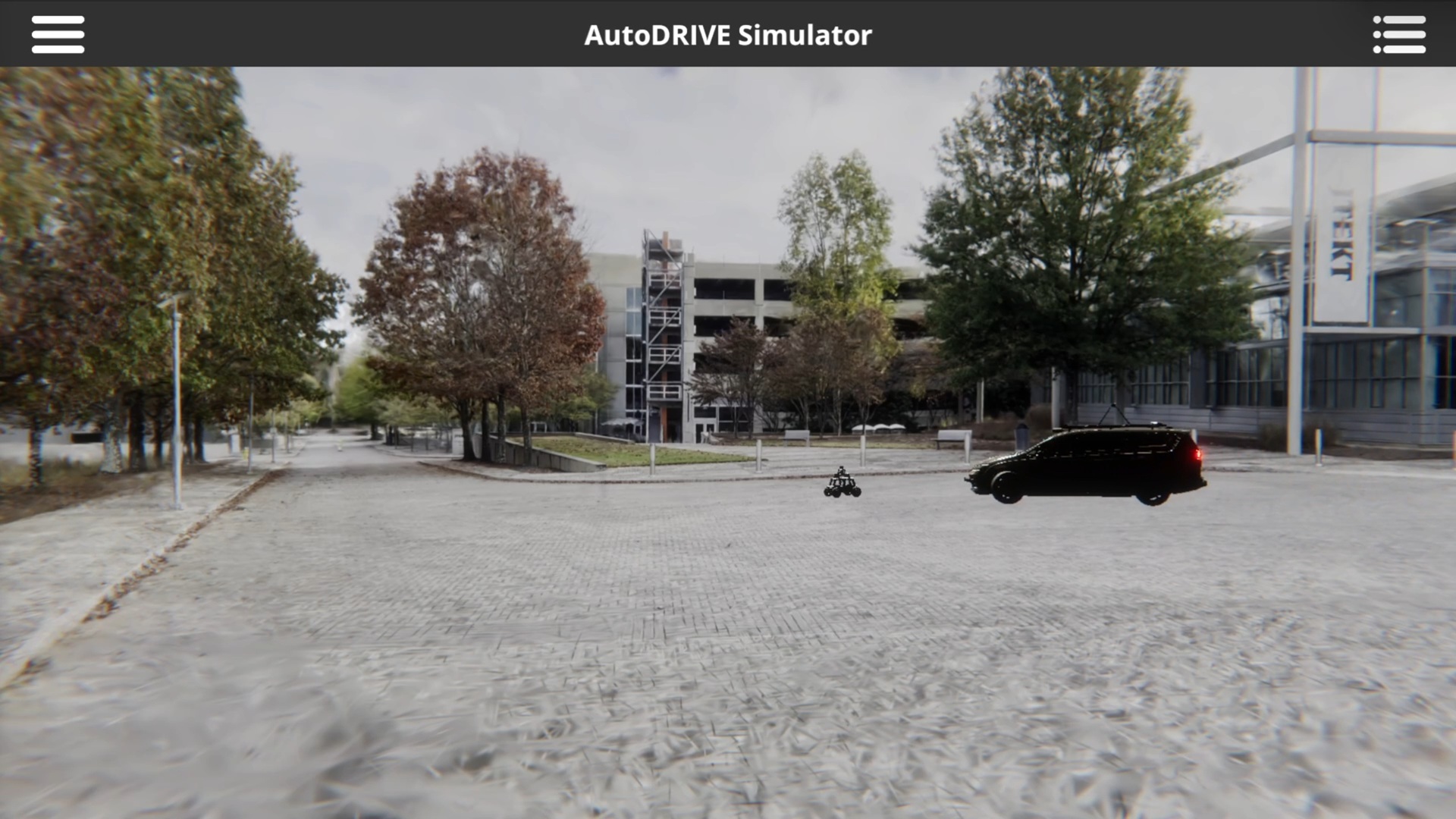}
         \caption{\textbf{Prompt:} Add a mid-scale Ackermann-steered robot performing skidpad maneuver.}
         \label{fig6k}
     \end{subfigure}
     \begin{subfigure}[b]{0.329\linewidth}
         \centering
         \includegraphics[width=\linewidth]{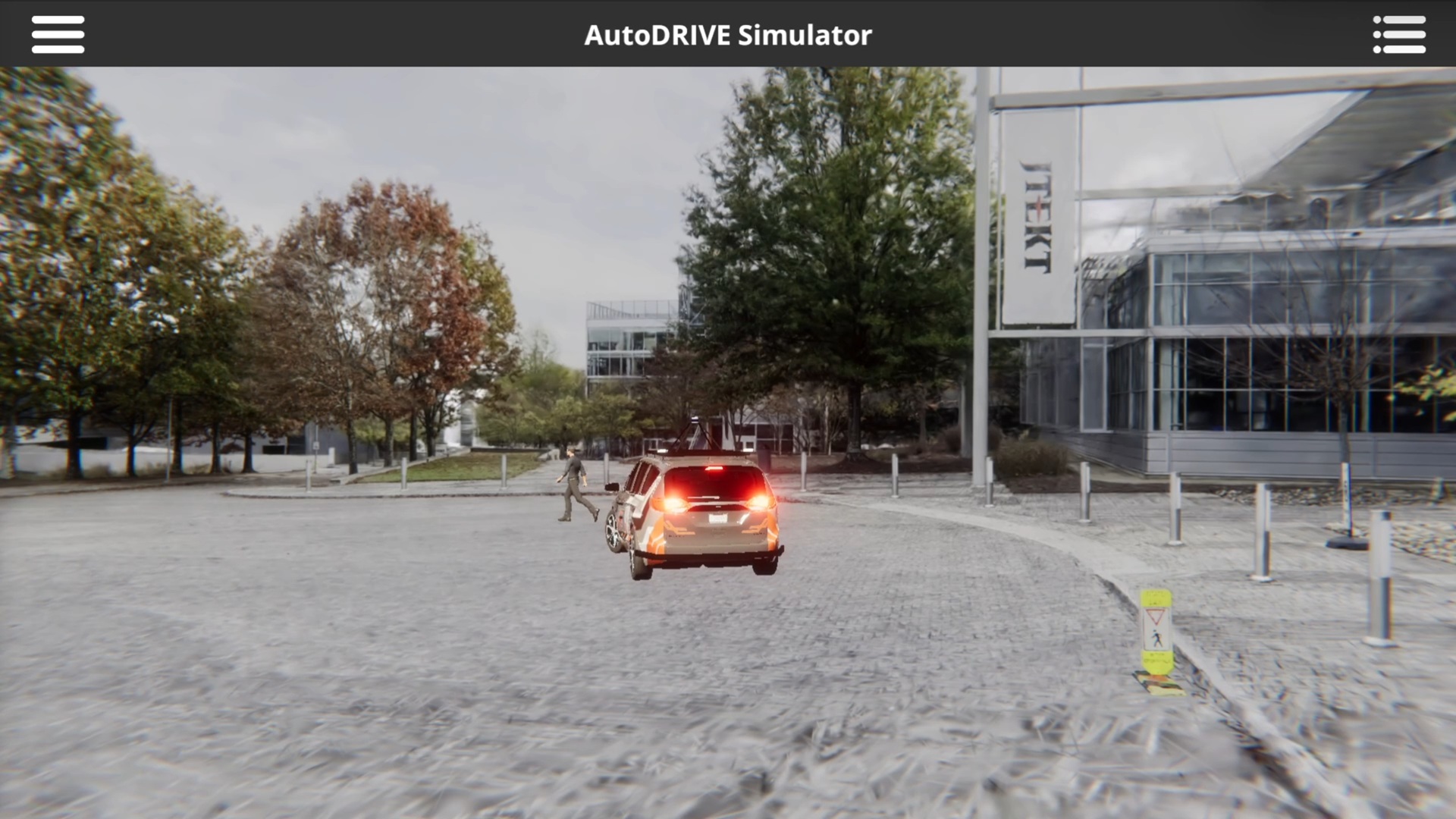}
         \caption{\textbf{Prompt:} Make a jaywalking pedestrian cut across the ego vehicle’s path.}
         \label{fig6l}
     \end{subfigure}
     \caption{LLM-guided reconfiguration: Digital twin of the CU-ICAR campus being reconfigured to (a) daytime; (b) nighttime; (c) foggy weather; (d) rainfall; (e) snowfall; (f) clear weather; (g-i) sequentially add maintenance work assets; (j) add a parked car at the specified location; (k) add a mobile robot controlled to perform a specific maneuver; (l) include a jaywalking pedestrian.}
    \label{fig6}
\end{figure*}

Following the benchmarking experiment, we analyzed the performance of 3D reconstruction. Here, we considered scene variations in terms of their setting (indoor/outdoor), scale (small/large), and lighting (lit/unlit). Diversity in assets was considered in terms of their color, texture, opacity, albedo, and scale. Table \ref{tab3} presents the data requirement, reconstruction quality, and FPS for 5 scenes and 3 assets depicted in Fig. \ref{fig4}.

It was observed that large outdoor scenes typically required upwards of 1k frames for effective reconstruction. Smaller (indoor) scenes, on the other hand, could be reconstructed with just half the data, but their depth estimates were slightly inaccurate, resulting in a few stray artifacts. This issue (inaccurate depth mapping leading to stray artifacts) was more pronounced in the footpath scenes, where data quantity was severely reduced. We also noticed that a well-lit footpath scene required significantly ($\sim$4$\times$) less data than a similar footpath scene, which was poorly lit.

Asset reconstructions also followed similar observations, where larger asset reconstructions were prone to reduced depth estimation accuracy, which could be attributed to data coverage limitations. For smaller assets, the data requirement was relatively less, but reflective (or retro-reflective) assets required slightly more data for improving depth estimation.

It was inferred that the scale or type of scene/asset did not directly govern the reconstruction fidelity. Instead, the quality (coverage, lighting, resolution, frame rate, etc.) of the data mattered most. However, it is also arguable that collecting good-quality data for smaller assets and indoor scenes is much easier than their larger and outdoor counterparts.

\begin{table}[t]
\centering
\caption{\small Scene Reconstruction Benchmarking}
\label{tab2}
\resizebox{\columnwidth}{!}{%
\begin{tabular}{l|l|l|l|l|l}
\hline
\textbf{Method} & \textbf{CCD/CMD $\downarrow$} & \textbf{PSNR $\uparrow$} & \textbf{SSIM $\uparrow$} & \textbf{LPIPS $\downarrow$} & \textbf{FPS $\uparrow$} \\ \hline
NeRF \cite{NeRF2021} & 1.83e-1 & 18.35 & 0.41 & 0.56 & 0.57   \\
3DGS \cite{kerbl20233dgaussiansplattingrealtime} & 6.54e-2 & \textbf{23.71} & \textbf{0.78} & \textbf{0.20} & 63.55  \\
PSR \cite{PoissonSurfaceRecon2006}  & \textbf{3.76e-4} & 8.74 & 0.32 & 0.89 & \textbf{144.26} \\
\textbf{3DGS+PSR (Ours)} & \textbf{3.76e-4} & \textbf{23.71} & \textbf{0.78} & \textbf{0.20} & 61.72  \\ \hline
\end{tabular}
}
\end{table}

\begin{table}[t]
\centering
\caption{\small Scene Reconstruction Performance Evaluation}
\label{tab3}
\resizebox{\columnwidth}{!}{%
\begin{tabular}{l|l|l|l|l|l}
\hline
\textbf{Scene/Asset} & \textbf{Data (\#)} & \textbf{PSNR $\uparrow$} & \textbf{SSIM $\uparrow$} & \textbf{LPIPS $\downarrow$} & \textbf{FPS $\uparrow$} \\ \hline
CU-ICAR              & 1420            	 & 23.71            	    & 0.78	                   & 0.20                        & 63.55 \\
CGEC                 & 1241            	 & 23.47            	    & 0.74	                   & 0.32                        & 61.96 \\
AuE Lab              & 545            	 & 35.10            	    & 0.97	                   & 0.17                        & 69.36 \\
Lit Footpath         & 70            	 & 32.20            	    & 0.93	                   & 0.09                        & 66.09 \\
Unlit Footpath       & 263            	 & 31.82            	    & 0.90	                   & 0.22                        & 68.27 \\
Passenger Car        & 971            	 & 24.93                    & 0.88                     & 0.19                        & 79.19 \\
Traffic Cone         & 530            	 & 35.11                    & 0.95                     & 0.06                        & 62.65 \\
Pedestrian Sign      & 1102            	 & 31.41	                & 0.94                     & 0.08                        & 64.40 \\ \hline
\end{tabular}
}
\end{table}

\subsection{Scenario Reconfiguration}

The design of experiments to analyze the LLM-guided scenario reconfiguration pipeline was split into benchmarking (refer Table \ref{tab4}) and evaluation (refer Table \ref{tab5}) categories.

In terms of benchmarking, we employed 3 different open-access LLMs, viz. Mistral 0.2, Llama 3.1, and Gemma 2. The choice of these models was primarily driven by their open accessibility, though it is worth mentioning that the proposed framework is capable of using any other models or APIs (both open as well as commercial). We consider the generalizability and repeatability of the model as our primary metrics for benchmarking, in addition to the model's disk space occupancy, number of parameters, and inference time (from prompt input to scenario update). Here, \textit{repeatability} is defined as the ability to produce the same result/response for 100 trials across 7 tasks with a direct prompt. Whereas, \textit{generalizability} is defined as the ability to reconfigure the scenario satisfactorily for 100 trials across 7 tasks with 4 grades of prompts.

The prompt types ranged across 4 different gradations:
\begin{itemize}
    \item \textbf{Direct Prompt:} These prompts directly express the user's intent, with exact asset names, numeric values, and other data as applicable. Such commands can be expected from domain experts.
    \item \textbf{Indirect Prompt:} These prompts clearly express the user's intent but do not provide exact asset names, numeric values, or other information. Such commands can be expected from users with a fair amount of knowledge, but who are not necessarily aware of specific terminology or conventions.
    \item \textbf{Vague Prompt:} These prompts express the user's intent minimally, and some things are left for interpretation by the LLM. Such commands can be expected from users with superficial or no knowledge.
    \item \textbf{Erroneous Prompt:} These prompts try to express the user's intent but include structural and/or grammatical errors. Such commands can be expected from non-native English speakers.
\end{itemize}

\begin{table}[t]
\centering
\caption{\small Scenario Reconfiguration Benchmarking}
\label{tab4}
\resizebox{\columnwidth}{!}{%
\begin{tabular}{l|l|l|l|l|l}
    \hline
    \textbf{LLM}  & \textbf{Size (GB)} & \textbf{Params. (\#)}  & \textbf{Gen. (\%) $\uparrow$}    & \textbf{Rep. (\%) $\uparrow$}    & \textbf{Time (s) $\downarrow$}   \\ \hline
    Mistral 0.2   & 4.06               & 7B                     & 46.21                 & 92.65                 & \textbf{6.60}                \\
    Llama 3.1     & 4.58               & 8B                     & 82.43                 & 89.38                 & 7.52                \\
    Gemma 2       & 5.36               & 9B                     & \textbf{84.93}                 & \textbf{94.86}                 & 8.45                \\ \hline
\end{tabular}
}
\end{table}

\begin{table}[t] 
\centering
\caption{\small Scenario Reconfiguration Performance Evaluation}
\label{tab5}
\resizebox{\columnwidth}{!}{%
\begin{tabular}{l|l|l|l|l}
    \hline
    \textbf{Task/Satisfaction}  &  \textbf{Direct Prompt}  & \textbf{Indirect Prompt}  &  \textbf{Vague Prompt}  &  \textbf{Erroneous Prompt} \\ \hline
    Search         &  99\%                    & 98\%                      &  80\%                   &  69\%                      \\
    Addition       &  99\%                    & 98\%                      &  90\%                   &  82\%                      \\
    Removal        &  100\%                   & 99\%                      &  88\%                   &  81\%                      \\
    Positioning    &  92\%                    & 86\%                      &  72\%                   &  65\%                      \\
    Moving         &  90\%                    & 87\%                      &  81\%                   &  73\%                      \\
    Arrangement    &  86\%                    & 85\%                      &  72\%                   &  57\%                      \\
    Appearance     &  98\%                    & 97\%                      &  79\%                   &  75\%                      \\ \hline
\end{tabular}
}
\end{table}

The tasks included (i) searching for specific assets or properties, (ii) addition of assets, (iii) removal of assets, (iv) positioning the assets according to spatial cues, (v) moving the assets, (vi) arranging homogeneous or heterogeneous assets, and (vii) modifying appearance of the scene/assets.

The reconfiguration capabilities of various LLMs were assessed based on whether they satisfactorily accomplished the intended task (a single task at a time). These assessments were based on the updated simulation states after scenario reconfiguration, which were manually analyzed against the intended task stated in the prompt (incorrect/inaccurate execution of the task or inability to comprehend the prompt were both considered failures).

In terms of model-specific performance (refer Table \ref{tab4}), it was observed that Mistral 0.2 provided highly overconfident responses, resulting in high repeatability without much generalizability. Llama 3.1 and Gemma 2 performed comparably, but the latter was slightly better and generated more elegant responses. Consequently, subsequent results will assume the Gemma 2 model, unless mentioned otherwise.

A detailed performance evaluation of the Gemma 2 model (refer Table \ref{tab5}) indicated that direct prompts resulted in over 90\% satisfaction for the majority of the tasks, except for arrangement, which was an inherently complex task. Reducing the quality of prompts to indirect structuring resulted in degradation of positioning and moving tasks, without significantly affecting others. Further degrading the prompt quality to vague structuring resulted in a significant degradation across all the tasks, with search and appearance tasks being affected the most. Finally, adulterating the prompts with errors resulted in a noticeable degradation across all the tasks, with arrangement and search tasks being affected the most.

\begin{figure*}[t]
     \centering
     \begin{subfigure}[b]{0.245\linewidth}
         \centering
         \includegraphics[width=\linewidth]{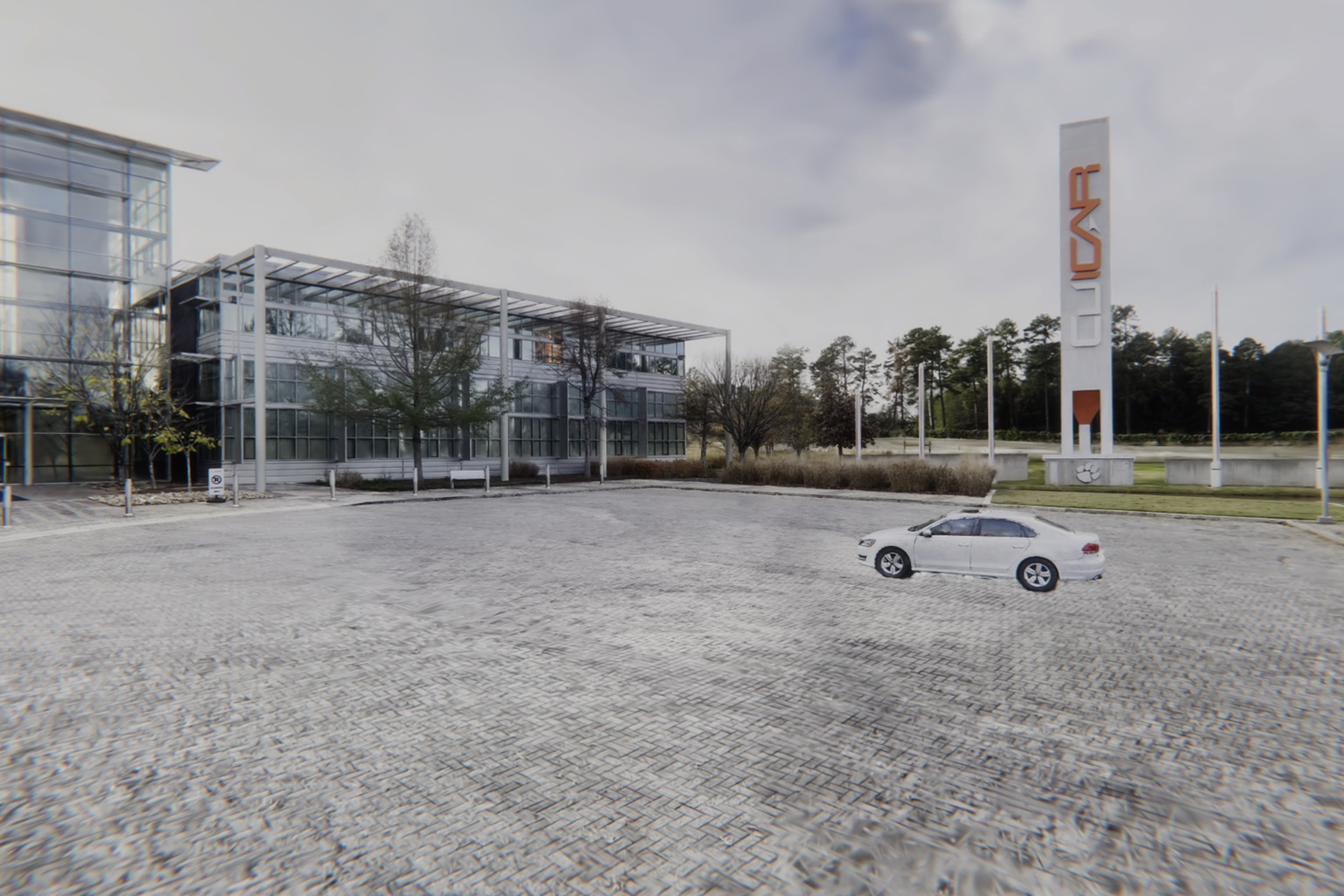}
         \caption{\textbf{Day:} Baseline}
         \label{fig7a}
     \end{subfigure}
     \begin{subfigure}[b]{0.245\linewidth}
         \centering
         \includegraphics[width=\linewidth]{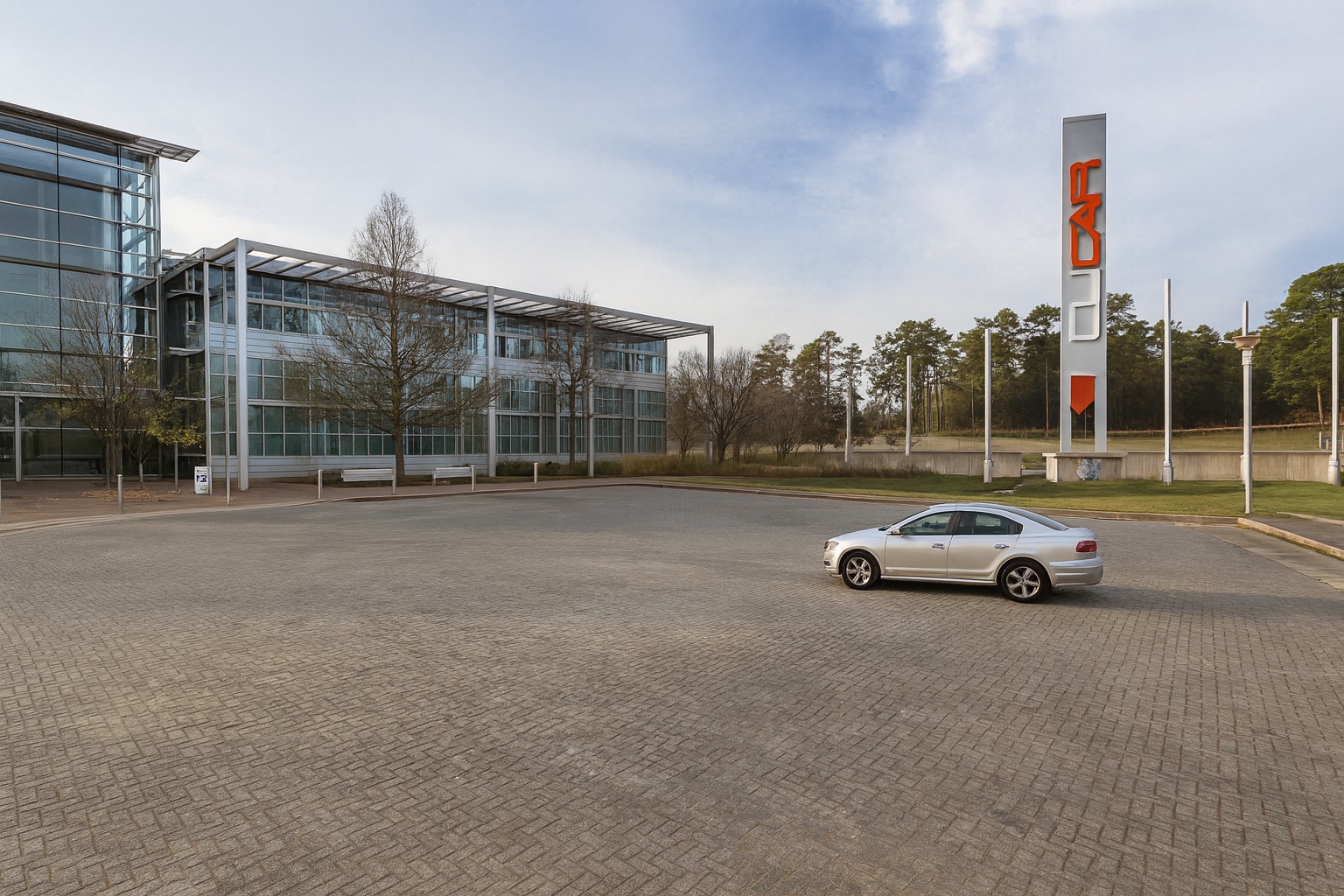}
         \caption{\textbf{Day:} Enhanced}
         \label{fig7b}
     \end{subfigure}
     \begin{subfigure}[b]{0.245\linewidth}
         \centering
         \includegraphics[width=\linewidth]{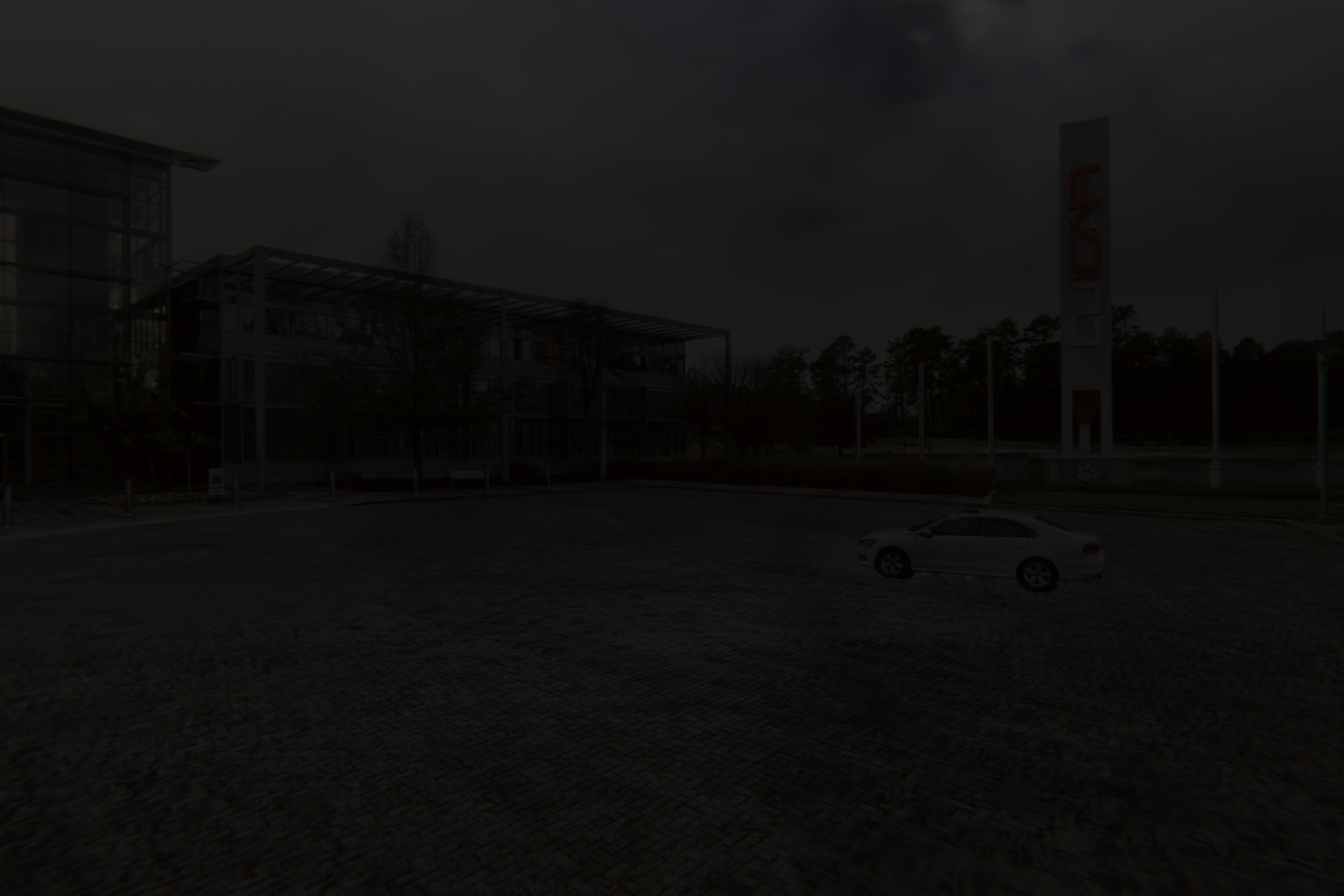}
         \caption{\textbf{Night:} Baseline}
         \label{fig7c}
     \end{subfigure}
     \begin{subfigure}[b]{0.245\linewidth}
         \centering
         \includegraphics[width=\linewidth]{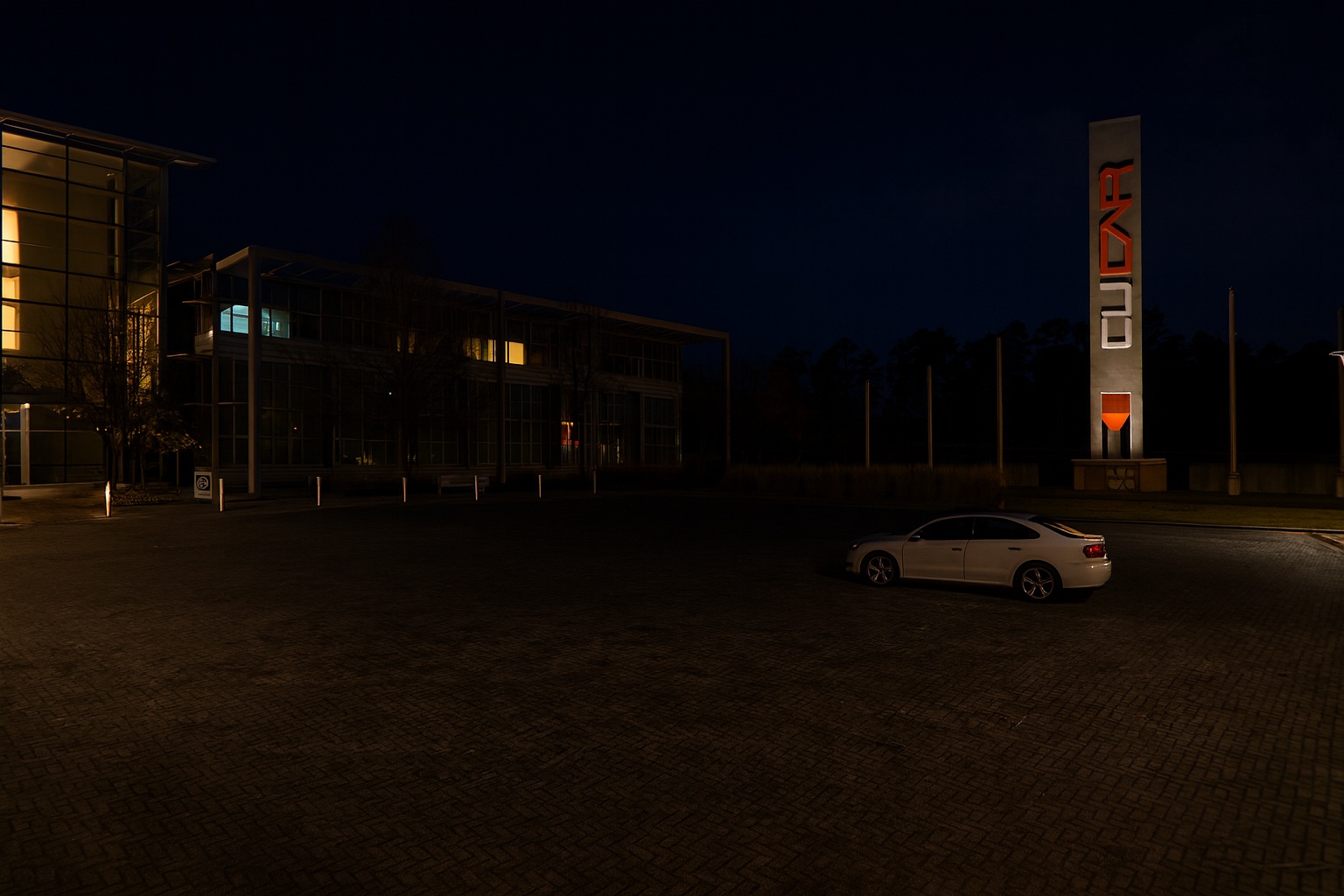}
         \caption{\textbf{Night:} Enhanced}
         \label{fig7d}
     \end{subfigure}
     \begin{subfigure}[b]{0.245\linewidth}
         \centering
         \includegraphics[width=\linewidth]{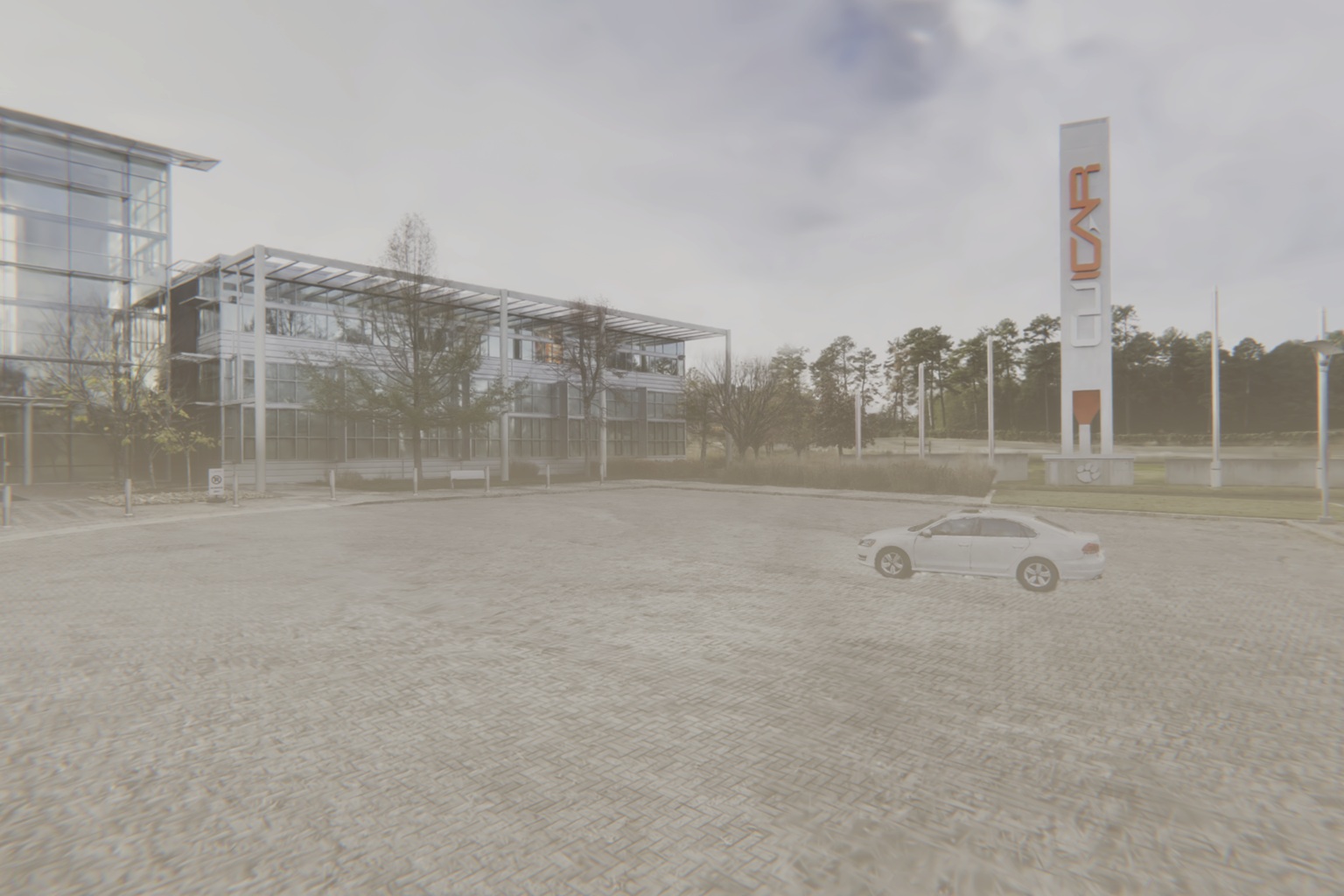}
         \caption{\textbf{Fog:} Baseline}
         \label{fig7e}
     \end{subfigure}
     \begin{subfigure}[b]{0.245\linewidth}
         \centering
         \includegraphics[width=\linewidth]{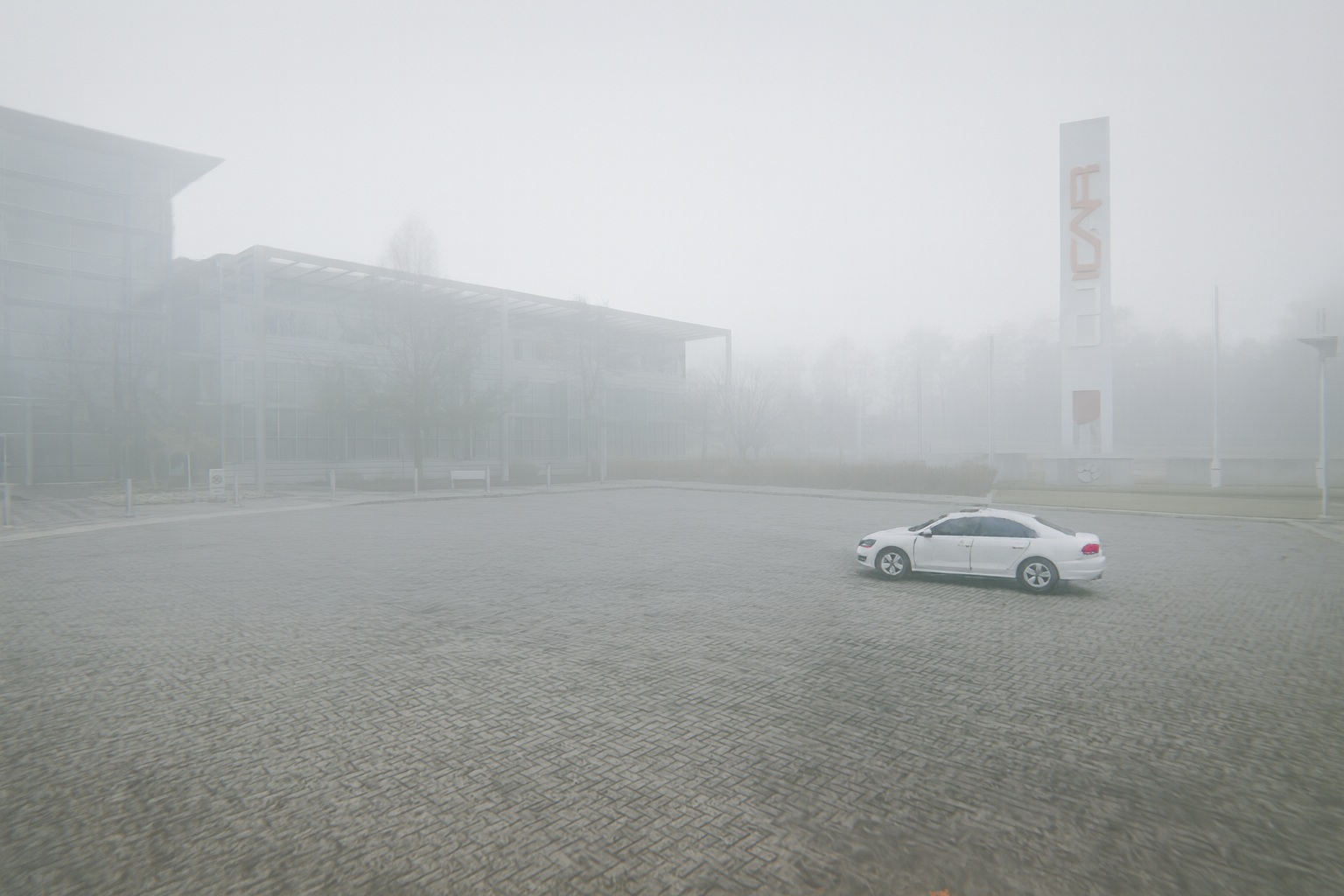}
         \caption{\textbf{Fog:} Enhanced}
         \label{fig7f}
     \end{subfigure}
     \begin{subfigure}[b]{0.245\linewidth}
         \centering
         \includegraphics[width=\linewidth]{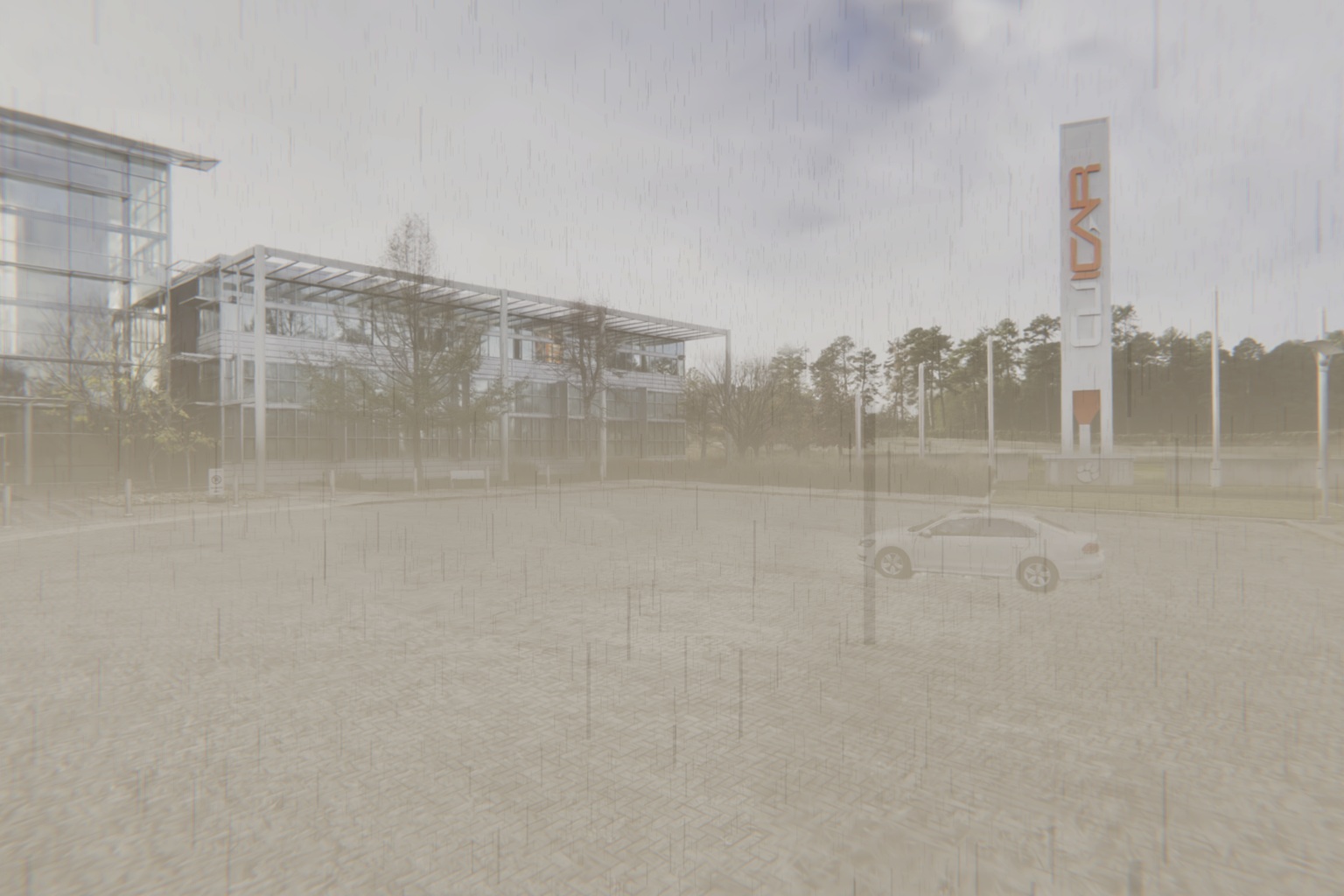}
         \caption{\textbf{Rain:} Baseline}
         \label{fig7g}
     \end{subfigure}
     \begin{subfigure}[b]{0.245\linewidth}
         \centering
         \includegraphics[width=\linewidth]{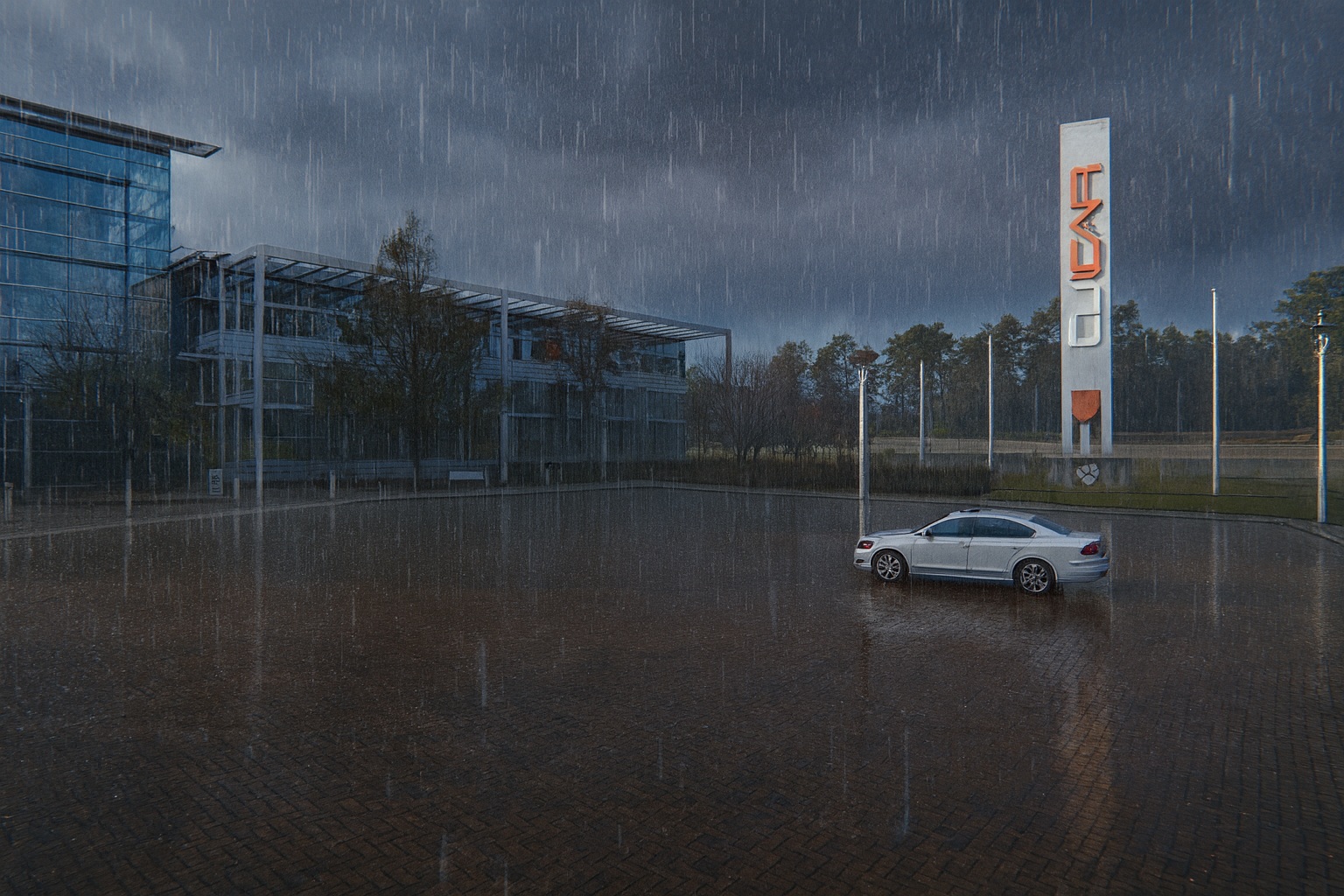}
         \caption{\textbf{Rain:} Enhanced}
         \label{fig7h}
     \end{subfigure}
     \begin{subfigure}[b]{0.245\linewidth}
         \centering
         \includegraphics[width=\linewidth]{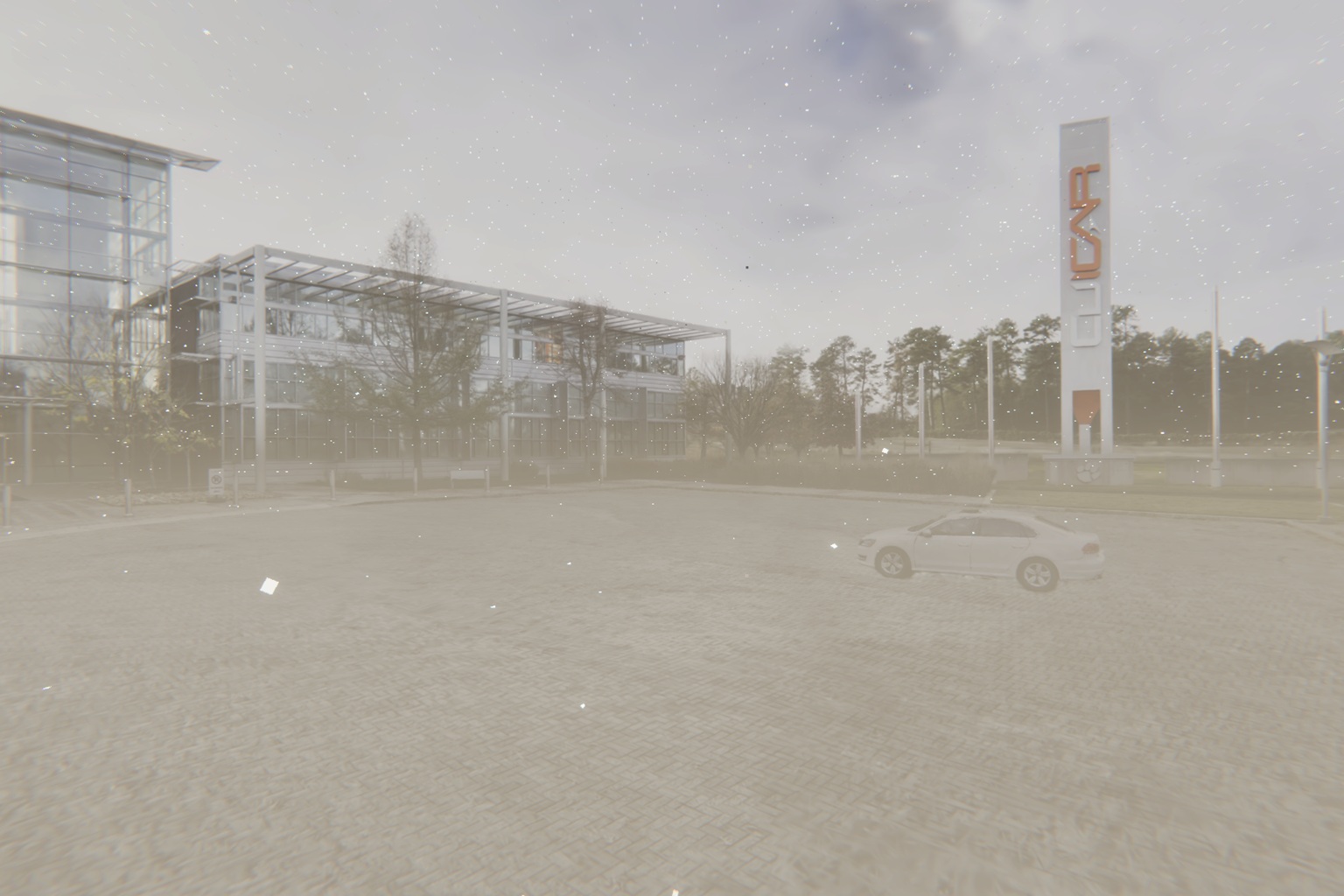}
         \caption{\textbf{Snow:} Baseline}
         \label{fig7i}
     \end{subfigure}
     \begin{subfigure}[b]{0.245\linewidth}
         \centering
         \includegraphics[width=\linewidth]{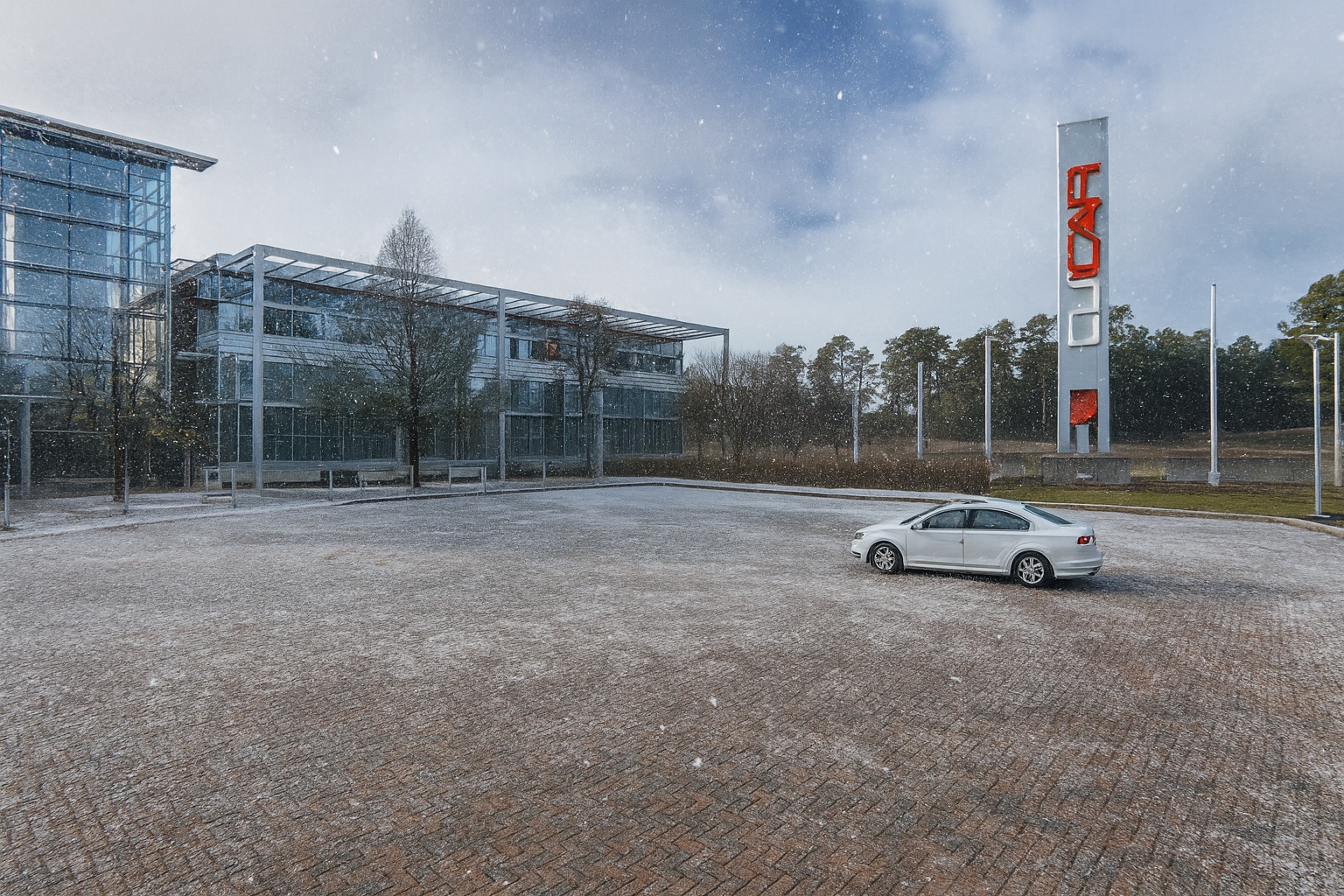}
         \caption{\textbf{Snow:} Enhanced}
         \label{fig7j}
     \end{subfigure}
     \begin{subfigure}[b]{0.245\linewidth}
         \centering
         \includegraphics[width=\linewidth]{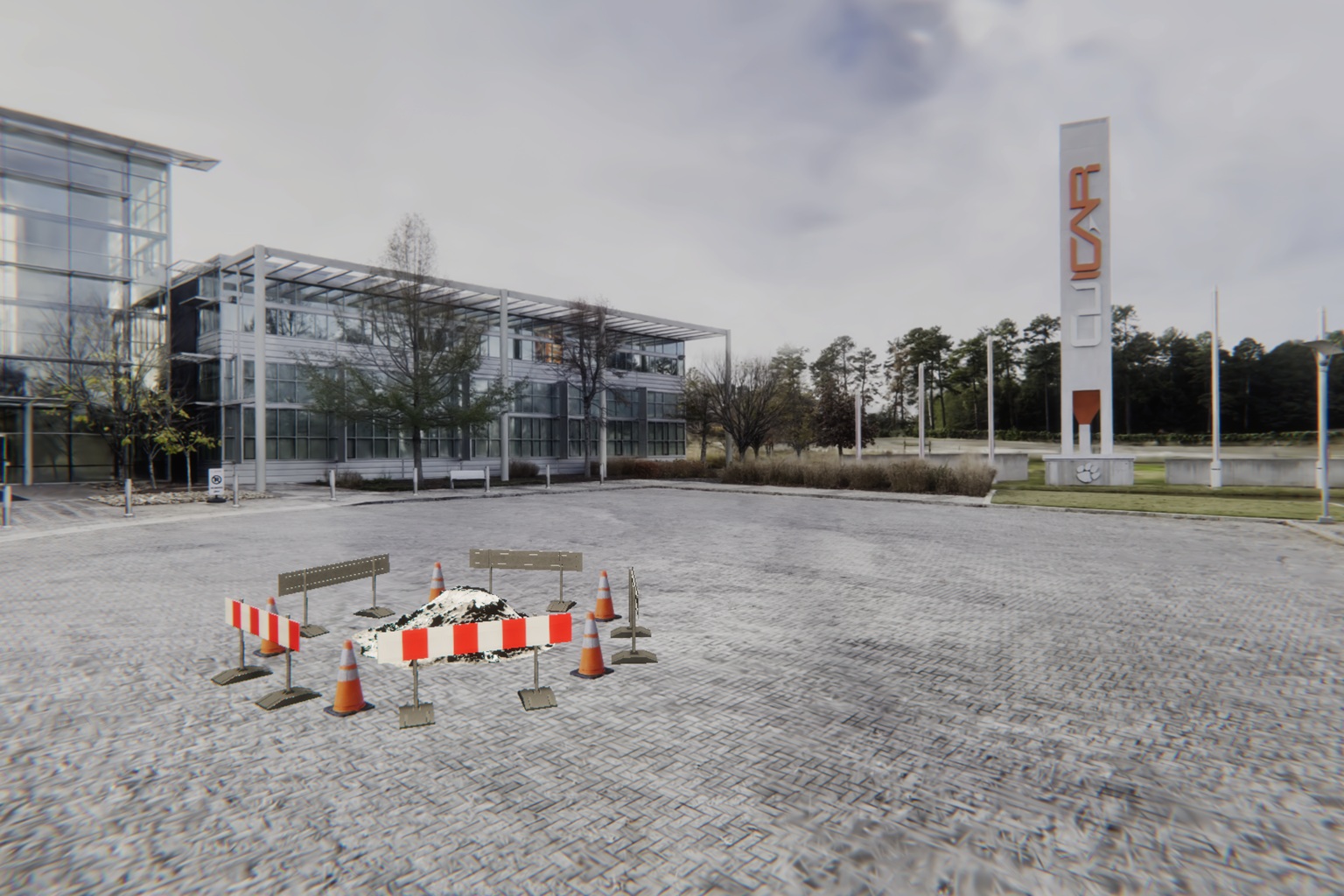}
         \caption{\textbf{Maintenance:} Baseline}
         \label{fig7k}
     \end{subfigure}
     \begin{subfigure}[b]{0.245\linewidth}
         \centering
         \includegraphics[width=\linewidth]{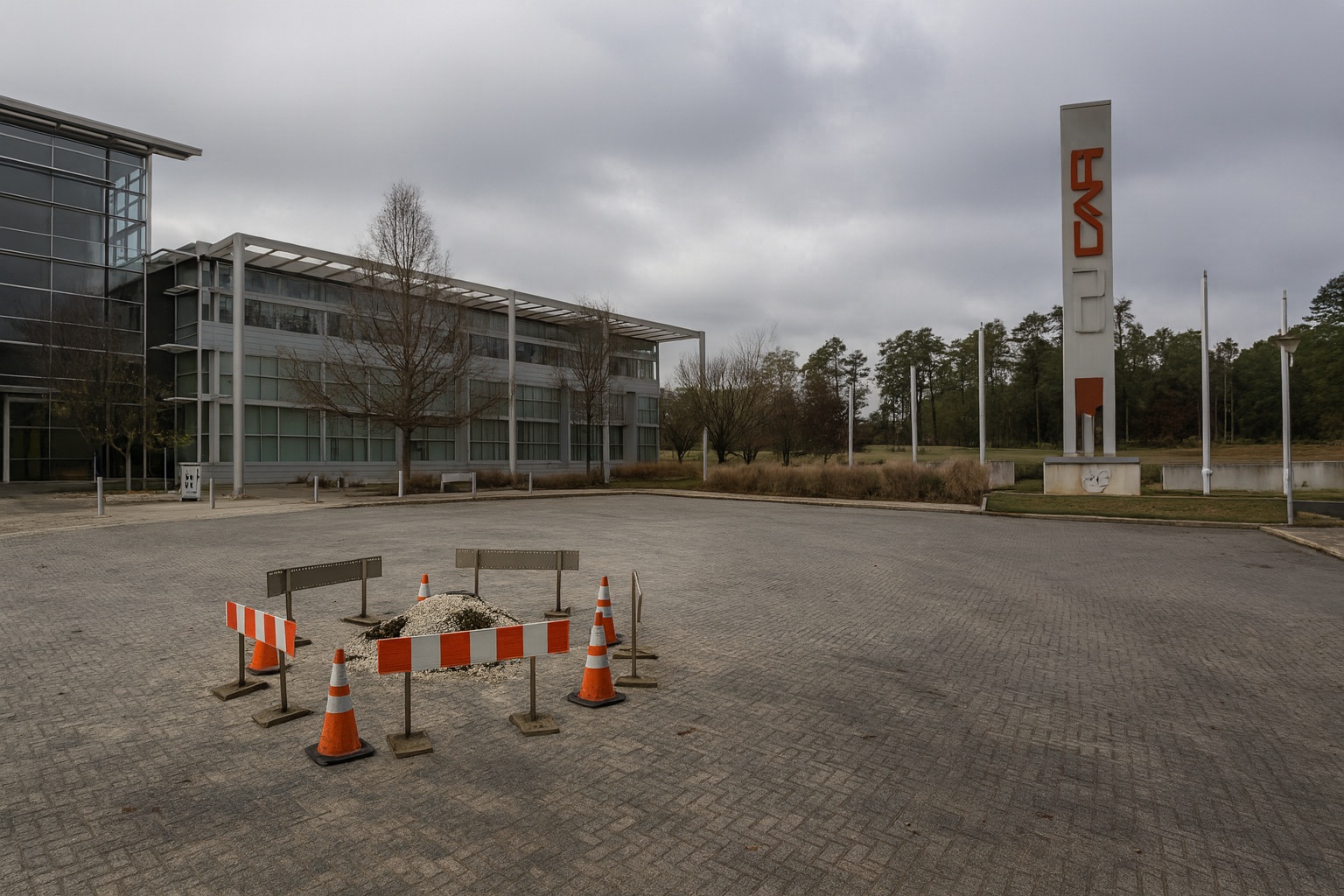}
         \caption{\textbf{Maintenance:} Enhanced}
         \label{fig7l}
     \end{subfigure}
     \caption{VLM-guided enhancement: Digital twin of the CU-ICAR campus being visually enhanced across (a-b) daytime; (c-d) nighttime; (e-f) foggy weather; (g-h) rainfall; (i-j) snowfall; (k-l) maintenance work. Notice appearance enhancement, including shading, lighting, reflections, shadows, and blending of volumetric effects, without affecting the pose or scale of the scene/assets.}
    \label{fig7}
\end{figure*}

We also conducted experiments with varying levels of prompt complexity and qualitatively analyzed them. These started with simple tasks like changing the time of day (refer Fig. \hyperref[fig6]{\ref*{fig6}(a-b)}) or weather conditions (refer Fig. \hyperref[fig6]{\ref*{fig6}(c-f)}). Next, we performed multi-round scenario editing using sequential prompting (refer Fig. \hyperref[fig6]{\ref*{fig6}(g-i)}) without repeating earlier information, to assess the context-retention ability. Finally, we also prompted complex commands involving several types of tasks to be executed simultaneously (refer Fig. \hyperref[fig6]{\ref*{fig6}(j-l)}) to assess the task management ability. It was interesting to see the spatial awareness of the model (e.g., implicit parking location in Fig. \hyperref[fig6]{\ref*{fig6}(j)}), as well as its context-aware creativity (e.g., adding a pedestrian sign without being explicitly asked in Fig. \hyperref[fig6]{\ref*{fig6}(l)}).

In terms of model size, parameters, and inference time, it was observed that for every 1B additional parameters, the model size increased by $\sim$0.5 GB, and its inference time by $\sim$1 second. However, for the objective of scenario reconfiguration, each increment in the parameter space significantly improved model performance (generalizability and repeatability). This performance improvement outweighed the slight increase in model size, given modern digital storage systems. Similarly, slightly higher inference time (order of milliseconds) could not be compared to a manual hand-crafting (order of hours/days).

\begin{table}[t]
\centering
\caption{Visual Enhancement Benchmarking}
\label{tab6}
\resizebox{\columnwidth}{!}{%
\begin{tabular}{l|l|l|l|l|l}
\hline
\textbf{Method} & \textbf{FID $\downarrow$} & \textbf{KID $\downarrow$} & \textbf{CMMD $\downarrow$} & \textbf{AIDP $\downarrow$} & \textbf{FPS $\uparrow$} \\ \hline
Baseline  & 13.33  & 0.44  & 0.57  & 0.55  & 63.55  \\
Enhanced  & 13.25  & 0.39  & 0.48  & 0.30  & 06.68  \\ \hline
\end{tabular}%
}
\end{table}

Ultimately, we benchmarked the optional VLM-based visual enhancement module. Particularly, we compared the baseline image frames with their enhanced counterparts qualitatively (refer Fig. \ref{fig7} for some examples) as well as quantitatively (refer Table \ref{tab6}) across the following metrics \cite{Stocco2024}:

\textbf{FID:} Fréchet Inception distance (FID) evaluates the realism of generated images by comparing the distributions of features extracted from generated and real images using a pretrained Inception network. It measures the distance between two multivariate Gaussian distributions characterized by their means and covariances. Lower FID indicates closer alignment to real image statistics (higher realism). While FID captures high-level semantic and perceptual differences better than pixel-wise metrics, it can be sensitive to sample size and feature extractor bias.

\textbf{KID:} Kernel Inception distance (KID) measures the similarity between generated and real image distributions using a kernel-based maximum mean discrepancy (MMD) computed on Inception features. Unlike FID, KID provides an unbiased estimator for finite sample sizes. Lower KID values indicate better agreement between generated and real image distributions. KID captures semantic discrepancies but, like FID, does not directly assess image fidelity at the pixel level.

\textbf{CMMD:} CLIP-MMD (CMMD) quantifies the alignment between generated and reference image distributions by computing the MMD in a shared CLIP embedding space. By leveraging CLIP’s multi-modal representations, this metric captures semantic and perceptual similarity that aligns well with human judgment. Lower CMMD indicates closer distributional similarity, though results depend on the chosen CLIP model and may overlook fine-grained visual artifacts.

\textbf{AIDP:} AI detector probability (AIDP) estimates the likelihood that an image is generated by an artificial model, typically produced by a trained classifier distinguishing real versus synthetic content. Lower probabilities suggest more realistic images, while higher probabilities indicate detectable generative artifacts. Although helpful in assessing detectability, these scores are model-dependent, can change over time, and are proxy measures of visual quality or semantic correctness.

Results consistently indicated lower AIDP values for the enhanced images, indicating higher visual realism or quality compared to their baseline counterparts. Furthermore, the FID, KID, and CMMD proxy metrics also confirmed this trend, indicating that the enhanced image distribution was closer to the distribution of \textit{``real''} images. However, it is worth mentioning that due to the added latency of the VLM inference, the simulation FPS dropped by an order of magnitude (from 63.55 to 6.68), though this bottleneck can be addressed relatively easily by leveraging high-performance computing (HPC) resources (e.g., $>$65 FPS with L40S GPU).


\section{Conclusion}
\label{Section: Conclusion}

This paper presented a novel, unified simulation framework designed to meet the multifaceted demands of autonomous driving research by combining dynamical fidelity, photorealistic rendering, context-aware scenario orchestration, and real-time performance. We demonstrated how the proposed framework effectively bridges the gap between real2sim fidelity and serviceability in digital twin simulations by integrating physics-based modeling with data-driven reconstruction, and enabling natural language interaction through an LLM interface. Our experiments indicated high levels of structural similarity ($\sim$97\%) in reconstructed scenes/assets, strong generalizability ($\sim$85\%) and repeatability ($\sim$95\%) in scenario reconfiguration, and real-time performance ($>$60 FPS) on a consumer-grade PC. Furthermore, the optional VLM-based image enhancement module consistently improved the output image quality ($\sim$80\%) by seamlessly blending the scene composition. These findings suggest that the proposed framework can serve as a scalable tool for advancing the development, optimization, and validation of autonomous driving systems under realistic and diverse conditions.

Looking ahead, we wish to address some of the limitations of the existing pipeline: (a) 3DGS does not fully support raytracing, which limits lighting effects such as shadows, reflections, etc.; (b) the demonstrated LLMs tend to get stuck at certain prompts without retaining the keywords and require specific prompt engineering to circumvent this issue, which cannot be guaranteed to work every time; and (c) The VLM-based enhancement pipeline is susceptible to generating varied outputs that may, occasionally, not completely adhere to the specified constraints.


\balance
\bibliographystyle{IEEEtran}
\bibliography{References}


\begin{IEEEbiography}[{\includegraphics[width=1in,height=1.25in,clip,keepaspectratio]{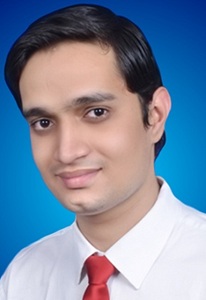}}]{Tanmay Samak} (Student Member, IEEE) received the B.Tech. degree in mechatronics engineering from SRM Institute of Science and Technology with a silver medal in 2021. He is currently pursuing a direct Ph.D. degree at Clemson University International Center for Automotive Research (CU-ICAR), where he is a member of the ARMLab. His research interests include autonomy-oriented modeling, estimation, and simulation methods aimed at bridging the real2sim gap for developing physically and graphically accurate digital twins.
\end{IEEEbiography}

\begin{IEEEbiography}[{\includegraphics[width=1in,height=1.25in,clip,keepaspectratio]{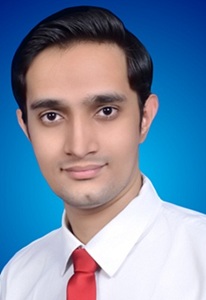}}]{Chinmay Samak} (Student Member, IEEE) received the B.Tech. degree in mechatronics engineering from SRM Institute of Science and Technology with a gold medal in 2021. He is currently pursuing a direct Ph.D. degree at Clemson University International Center for Automotive Research (CU-ICAR), where he is a member of the ARMLab. His research interests lie at the intersection of physics-informed and data-driven techniques for identification, adaptation, and augmentation aimed at bridging the sim2real gap using autonomy-oriented digital twins.
\end{IEEEbiography}

\begin{IEEEbiography}[{\includegraphics[width=1in,height=1.25in,clip,keepaspectratio]{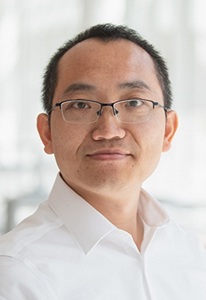}}]{Bing Li} (Member, IEEE) received the Ph.D. degree in electrical engineering from the City University of New York (CUNY) in 2018. He is an Associate Professor with the Department of Automotive Engineering at Clemson University, where he also directs the AutoAI Lab. The central focus of his research is spatial intelligence for safer/assistive mobility and robots in dynamic and interactive environments, including topics such as multi-modal perception, 3D reconstruction and SLAM, deep learning, autonomous agents, and human-centered AI.
\end{IEEEbiography}

\begin{IEEEbiography}[{\includegraphics[width=1in,height=1.25in,clip,keepaspectratio]{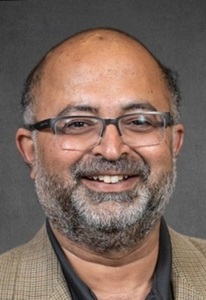}}]{Venkat Krovi} (Senior Member, IEEE) received the Ph.D. degree in mechanical engineering and applied mechanics from the University of Pennsylvania in 1998. He is the Michelin Endowed Chair Professor of Vehicle Automation with the Departments of Automotive and Mechanical Engineering at Clemson University, where he also directs the ARMLab. The underlying theme of his research has been to take advantage of ``power of the many'' (both autonomous agents and humans) to extend the reach of human users into dull, dirty, and dangerous environments.
\end{IEEEbiography}

\vfill

\end{document}